%% file: main.tex
\documentclass[10pt,twocolumn,letterpaper]{article}

\usepackage{cvpr}              
\usepackage{url}
\usepackage{xurl}
\usepackage{amsmath}
\usepackage{multicol}
\usepackage{multirow}
\usepackage{booktabs}
\usepackage{bbding}
\usepackage{graphicx}
\usepackage{float}
\usepackage{stfloats}
\usepackage{pifont}
\usepackage{makecell}
\usepackage{microtype}

\input{preamble}
\definecolor{cvprblue}{rgb}{0.21,0.49,0.74}
\usepackage[pagebackref,breaklinks,colorlinks,allcolors=cvprblue]{hyperref}

\def\paperID{10353} 
\def\confName{CVPR}
\def\confYear{2026}

\title{4DEquine: Disentangling Motion and Appearance for 4D Equine Reconstruction from Monocular Video}

\author{ Jin Lyu$^{1}$, Liang An$^{2,\dag}$, Pujin Cheng$^{1,3}$, Yebin Liu$^{2,\dag}$, Xiaoying Tang$^{1,\dag}$\\
$^{1}$Southern University of Science and Technology
$^{2}$Tsinghua University 
$^{3}$The University of Hong Kong
}

\begin{document}
\maketitle
\input{sec/00_abstract}
\input{sec/01_introduction}
\input{sec/02_related_works}
\input{sec/03_preliminary}
\input{sec/04_method}
\input{sec/05_experiment}
\input{sec/06_conclusion}
{
    \small
    \bibliographystyle{ieeenat_fullname}
    \bibliography{main}
}
\input{sec/X_suppl}


\end{document}

%% file: sec/00_abstract.tex
\begin{abstract}
4D reconstruction of equine family (e.g. horses) from monocular video is important for animal welfare. Previous mainstream 4D animal reconstruction methods require joint optimization of motion and appearance over a whole video, which is time-consuming and sensitive to incomplete observation. In this work, we propose a novel framework called 4DEquine by disentangling the 4D reconstruction problem into two sub-problems: dynamic motion reconstruction and static appearance reconstruction. For motion, we introduce a simple yet effective spatio-temporal transformer with a post-optimization stage to regress smooth and pixel-aligned pose and shape sequences from video. For appearance, we design a novel feed-forward network that reconstructs a high-fidelity, animatable 3D Gaussian avatar from as few as a single image. To assist training, we create a large-scale synthetic motion dataset, VarenPoser, which features high-quality surface motions and diverse camera trajectories, as well as a synthetic appearance dataset, VarenTex, comprising realistic multi-view images generated through multi-view diffusion. While training only on synthetic datasets, 4DEquine achieves state-of-the-art performance on real-world APT36K and AiM datasets, demonstrating the superiority of 4DEquine and our new datasets for both geometry and appearance reconstruction. Comprehensive ablation studies validate the effectiveness of both the motion and appearance reconstruction network. Project page: \url{https://luoxue-star.github.io/4DEquine_Project_Page/}. 
\end{abstract}

%% file: sec/01_introduction.tex
\section{Introduction}
Equines, which include horses, donkeys, and zebras, have played a crucial role in human civilization. They serve as a core driving force for transportation, agriculture, and warfare, extensively promoting social and historical progress. To this day, they continue to contribute to human welfare in areas such as livestock farming, sports and athletics, cultural symbolism, and even animal-assisted therapy.
Monocular 4D reconstruction of equines aims to realistically recover their shape, motion, and appearance from a convenient, single-view video input, producing valuable digital copies of real-world equine activities.

Previously, 
monocular 4D reconstruction of \textbf{general} scenes can reconstruct dynamic objects and static backgrounds from video~\citep{feng2025st4rtrack,wang2025shape,wu20254d}, but they fail to recover the full geometry of animals from incomplete observations. 
For the \textbf{specific} task of animal reconstruction, several approaches build animatable avatars from videos or casual images, yielding per-frame 3D (i.e. 4D) representations. 
Template-free approaches, which estimate deformable radiance fields~\citep{yang2022banmo,yang2023reconstructing} or point clouds~\citep{kaye2025dualpm}, often lack an explicit structural prior, degrading their geometric detail. 
Conversely, methods leveraging the 3D statistical mesh model SMAL~\citep{Zuffi:CVPR:2017} (e.g., SMALR~\citep{zuffi2018lions} and SMALST~\citep{Zuffi:ICCV:2019}) provide a strong geometric prior but they extract textures directly from the input image, making the final appearance highly sensitive to mesh-to-image alignment accuracy.
More advanced techniques replace textures with powerful representations like implicit duplex fields~\citep{AnimalAvatars2024,zhong20254d} or mesh-bound Gaussians~\citep{lei2024gart, cho2025dogrecon}.
However, these methods are optimization-based, creating a critical bottleneck: they are computationally expensive and demand clean, 360-degree ``turntable'' sequences, which are rare in real-world recordings .

To overcome the drawbacks of optimization-based methods, we propose \textbf{4DEquine}. Our key idea is to explicitly disentangle the challenging 4D reconstruction problem into two sub-problems: recovering the dynamic motion (i.e., the motion sequence) and reconstructing the high-fidelity static appearance (i.e., a canonical 3D Gaussian avatar). The bridge that connects motion and appearance is the VAREN model~\cite{Zuffi:CVPR:2024}, a high-quality mesh model specifically designed for horses. Given a video sequence, 4DEquine recovers the accurate motion for each frame and generates Gaussian point clouds bound with VAREN mesh faces in a feed-forward manner. Achieving this goal requires novel network architectures and 4D data, which are our key contributions. 

Specifically, we first introduce \textbf{AniMoFormer}, a spatio-temporal transformer network combined with a Post-Optimization step to recover accurate and smooth motion from video clips. By processing consecutive frames within a time window, AniMoFormer outputs temporally consistent parameters, overcoming the jittering inherent in previous single-image methods~\cite{niewiadomski2025ICCV,lyu2025animer}. Training such a network, however, requires high-quality video data with 4D VAREN annotations, which are not publicly available. To fill this gap, we create VarenPoser, the first large-scale synthetic horse dataset with 4D VAREN annotations. VarenPoser's ground truth is generated by fitting VAREN meshes to the marker-based PFERD~\cite{li2024poses} motion dataset and texturing them using MV-Adapter~\cite{huang2024mvadapter}. A key innovation of VarenPoser is the introduction of diverse virtual camera trajectories to simulate real-captured camera movements during rendering. By training on VarenPoser only, AniMoFormer yields realistic horse motions on real-world data. Moreover, the post-optimization step further ensures pixel-aligned equine geometry recovery. 

Second, we propose \textbf{EquineGS}, a novel feed-forward framework that reconstructs a high-fidelity, animatable 3D Gaussian avatar from just a single, representative image. The observations behind single image feed-forward design are twofold. On the one hand, the appearance of an equine animal usually remains unchanged within the same video clip. On the other hand, most real-world videos cannot capture all viewpoints of an animal. Consequently, EquineGS enables rapid and high-fidelity horse appearance inference in the form of Gaussian point clouds.  
Note that this approach requires consistent multi-view data for training, which is hard to obtain in the real world. Therefore, we create VarenTex, another new synthetic dataset generated using a multi-view diffusion model, which provides higher-quality appearances than VarenPoser in the form of multi-view images. By training on VarenTex only, EquineGS directly predicts Gaussian primitives attached to the VAREN parametric model from real images, which are animated to each time point via linear blend skinning (LBS).

Finally, by integrating AniMoFormer and EquineGS, the full 4DEquine framework reconstructs 4D geometry and appearance of equines from monocular video, and could process videos of arbitrary length using sliding-window strategy. By testing on the equine subset of unseen video datasets APT-36K~\citep{yang2022apt} and AiM~\citep{Animal4D}, we achieve new state-of-the-art motion recovery performance on metrics like PCK@0.05, PCK@0.1, Accel and Chamfer Distance (CD). 
For full 4D reconstruction (geometry + appearance), 
4DEquine achieves superior 4D reconstruction quality on the AiM dataset compared to optimization-based methods GART~\citep{lei2024gart} and 4D-Fauna~\citep{Animal4D},
and outperforms recent general video-to-4D synthesis method GVFDiffusion~\citep{zhang2025gaussian}. 
Ablation studies demonstrate the effectiveness of our network design for both geometry and appearance reconstruction. 
In summary, 4DEquine provides an effective tool for the 4D reconstruction of equine animals, which will play an important role in downstream applications. 


%% file: sec/02_related_works.tex
\section{Related Works}
\noindent\textbf{4D Reconstruction Methods.}
General monocular 4D reconstruction methods aim to capture both static and dynamic scene geometries while building temporal motion correspondences~\citep{park2021nerfies, wang2023tracking, wu20244d, karaev2024cotracker, feng2025st4rtrack, wang2025shape, wu20254d}. A significant stream of this research requires training for each scene, which is time-consuming and fails to recover unseen surfaces. More recent feed-forward approaches, such as Monst3R~\citep{zhang2024monst3r} and Page-4D~\citep{zhou2025page}, leverage foundational visual geometry model~\citep{wang2025vggt} to eliminate this per-scene training requirement. However, a key limitation persists: both per-scene and generalizable methods still struggle to reconstruct the full geometry of an object from incomplete observations. For instance, reconstructing the full geometry and appearance of a horse using the above methods requires a 360-degree video recording around the target, which hinders real-world applications. 

\noindent\textbf{Equine Geometry Estimation.} 
Existing methods for 3D animal reconstruction are broadly categorized as model-free~\citep{kanazawa2018learning, yang2021lasr, yang2021viser, yang2022banmo,yang2023reconstructing, yao2022lassie, yao2023hi, wu2023dove, wu2023magicpony, li2024learning, Animal4D, jakab2024farm3d, kaye2025dualpm} or model-based methods; our work focuses on the latter. Since the proposal of the SMAL model~\citep{Zuffi:CVPR:2017}, equine animals (e.g., horses and zebras) have attracted great efforts for statistical modeling. SMALR~\citep{zuffi2018lions} uses shape deformation and texture stitching for general quadrupeds reconstruction from images, while SMALST~\citep{Zuffi:ICCV:2019} focuses on zebras in a similar way. hSMAL~\citep{li2021hsmal} and Dessie~\citep{li2024dessie} add horse-specific details to SMAL, yet still lack realism. A fundamental progress in horse modeling is the VAREN model~\citep{Zuffi:CVPR:2024}, which is learned from multiple high-resolution horse scans. Although many methods enable monocular horse geometry recovery by estimating the pose and shape parameters of SMAL from images such as SMALify~\citep{biggs2018creatures}, AniMer~\citep{lyu2025animer,lyu2025animerunifiedposeshape}, and GenZoo~\citep{niewiadomski2025ICCV}, none of them consider temporally 4D reconstruction, or make use of the state-of-the-art VAREN model. 

\noindent\textbf{Animal Avatar Reconstruction.}
One major line of work relies on per-instance optimization. Methods like BANMo~\citep{yang2022banmo} and RAC~\citep{yang2023reconstructing} learn deformable neural models from videos, but are often slow and lack fine geometric detail. Other optimization-based techniques reconstruct duplex-mesh avatars~\citep{AnimalAvatars2024, zhong20254d} or leverage 3D Gaussian Splatting (3DGS), such as GART~\citep{lei2024gart} and DogRecon~\citep{cho2025dogrecon}. However, these methods all require heavy per-instance optimization and enough coverage of observation perspectives. A second line of work focuses on scalable, feed-forward models. MagicPony~\citep{wu2023magicpony} and Ponymation~\citep{sun2024ponymation} learn animatable horses from casual images and videos. 3D-Fauna~\citep{li2024learning} learns a single pan-category model capable of reconstructing avatars for over 100 quadruped species from internet images, which is then extended to 4D-Fauna~\citep{Animal4D} on videos. Farm3D~\citep{jakab2024farm3d} shares similar principle yet different training strategy with 3D-Fauna. However, they must trade shape realism for diversity. 
Therefore, a feed-forward method for high-quality equine avatar generation is still lacking. 

%% file: sec/03_preliminary.tex
\section{Preliminary}
\noindent{\textbf{The VAREN Model.}} The VAREN (Very Accurate and Realistic Equine Network) model~\citep{Zuffi:CVPR:2024} is a data-driven, 3D parametric model of the horse, learned from thousands of 3D scans of 50 real horses of varying breeds and sizes. Unlike previous animal parametric models (e.g., SMAL~\citep{Zuffi:CVPR:2017}), the VAREN model introduces muscle deformation to model the non-rigid, pose-dependent deformations of the surface. Specifically, it groups body surface points into regions that correspond to the horse's superficial muscles and learns the influence of body part articulation on the observable deformation of each specific muscle from the scan data. Functionally, the model takes a set of input parameters—shape \(\boldsymbol{\beta} \in \mathbb{R}^{39}\), pose \(\boldsymbol{\theta} \in \mathbb{R}^{38 \times 3}\) (representing axis-angle rotations for 38 joints), and global translation \(\boldsymbol{\gamma} \in \mathbb{R}^{3}\) to generate a high-resolution 3D mesh with vertices \(V \in \mathbb{R}^{13873 \times 3}\) and faces \(F \in \mathbb{N}^{27706 \times 3}\) through shape blend shapes, muscle deformation, and a LBS process.

\noindent{\textbf{3D Gaussian Splatting (3DGS).}} 3DGS~\citep{kerbl3Dgaussians} models scenes as collections of 3D Gaussians. Each Gaussian primitive is defined by its position \(\boldsymbol{\mu}\ \in \mathbb{R}^{3}\), opacity value \(o \in [0, 1]\), scaling vector \(\boldsymbol{s} \in \mathbb{R}^{3}\), rotation quaternion \(\boldsymbol{r} \in \mathbb{R}^{4}\) and directional appearance \(\boldsymbol{f} \in \mathbb{R}^{C}\), which is represented by Spherical Harmonics (SH) coefficients. 
During rendering, these 3D Gaussian primitives are projected into 2D screen space for each camera view and then composited into the final image using a fast, tile-based rasterizer that performs view-ordered alpha-blending on depth-sorted primitives.

%% file: sec/04_method.tex
\section{Method}
Our framework, 4DEquine, consists of two disentangled components: AniMoFormer for spatio-temporal geometry recovery and EquineGS for feed-forward Gaussian avatar reconstruction, as illustrated in Fig.~\ref{fig:pipeline}. We detail the technical architectures and the corresponding datasets for each component Sec.~\ref{subsec:pose_shape_estimation} and Sec.~\ref{sec:avatar_reconstuction}, respectively. For more details, please refer to the supplementary materials.
\subsection{Temporal VAREN Recovery}
\label{subsec:pose_shape_estimation}
To address the scarcity of annotated 4D data, we construct \textbf{VarenPoser}, a large-scale synthetic video dataset to facilitate temporal training.
Then, we introduce \textbf{AniMoFormer}, a two-stage framework comprising a spatio-temporal transformer followed by a Post-Optimization stage. 
\subsubsection{VarenPoser Dataset Construction}
\begin{figure}[ht]
    \centering
    \includegraphics[width=\linewidth]{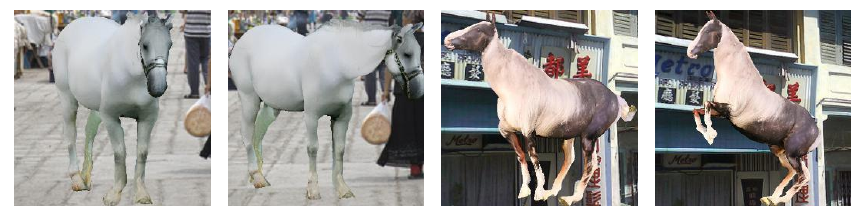}
    \caption{\textbf{Two samples of the VarenPoser dataset.} Each sample is a video, and we only show two frames for each sample. }
    \label{fig:varenposer}
\end{figure}

We build VarenPoser (samples in Fig.~\ref{fig:varenposer}) by fitting the VAREN model to the marker-based horse motion dataset PFERD~\cite{li2024poses} to acquire the pose parameters, which are then segmented into 600-frame clips.
Similar to Dessie~\citep{li2024dessie}, we assign a random shape parameter to each clip for morphological diversity. To generate varied yet temporally consistent appearances, we employ the MV-Adapter~\citep{huang2024mvadapter}, a texture generation model, to create diverse textures using a collection of horse images gathered from the internet. Unlike Dessie, which only creates static data, we render videos by simulating real-world camera movements. We implement three common yet distinct camera trajectories: fix, dolly, and orbit (see supplementary material for details).
The final VarenPoser dataset contains 1171 video clips, encompassing a wide variety of motions and appearances. All synthetic videos are with $512\times512$ resolution and 60 FPS. 

\subsubsection{AniMoFormer}
\begin{figure*}[ht]
    \centering
    \includegraphics[width=\linewidth]{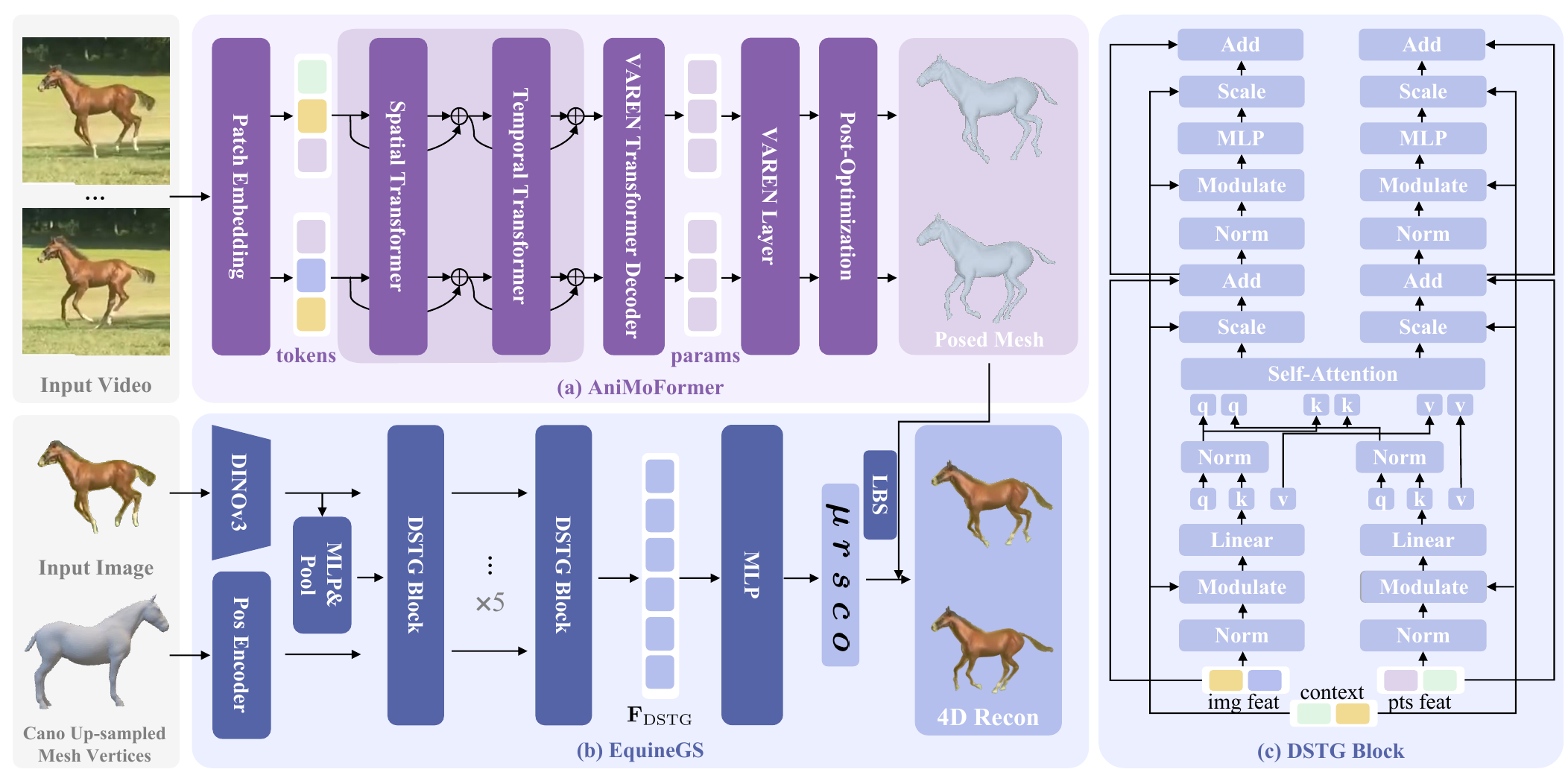}
    \caption{\textbf{Overview of 4DEquine.} 
    \textbf{(a) AniMoFormer:} A spatio-temporal transformer with post-optimization for motion recovery.
    \textbf{(b) EquineGS:} A feed-forward network to reconstruct a canonical 3D Gaussian avatar from a single image.
    \textbf{(c) DSTG-Block:} The dual-stream architecture used in EquineGS.
    }
    \label{fig:pipeline}
\end{figure*}

\noindent\textbf{Spatio-Temporal Transformer.}   
The first stage of AniMoFormer is a new spatio-temporal transformer that builds upon AniMer~\cite{lyu2025animer}. While AniMer estimates animal shape and pose from a single image using transformer, it can not leverage temporal information. 
To address this, and drawing inspiration from time-series modeling~\citep{Vaswani2017AttentionIA, yan2023temporally, li2024multi, wang2024tram}, our network is designed to process video clips. As illustrated in Fig.~\ref{fig:pipeline}, our model first processes each frame of an input sequence individually using a Spatial Transformer to extract a set of spatial features. 
A stack of N-frame spatial features is then fed into a Temporal Transformer, 
which uses self-attention across the entire N-frame window to model temporal relationships and capture local motion context. Finally, these motion-encoded features pass to a VAREN transformer decoder, which regresses the pose, shape, and camera parameters for the window. The training loss is as follows:
\begin{equation}
\begin{aligned}
    \label{animoformer_loss}
    &\mathcal{L} = \lambda_{\text{varen}} \mathcal{L}_{\text{varen}} + \lambda_{\text{smooth}} \mathcal{L}_{\text{smooth}} 
    +
    \lambda_{\text{2D}} \mathcal{L}_{\text{2D}} + 
    \lambda_{\text{3D}} \mathcal{L}_{\text{3D}}, \\
    &\mathcal{L}_\text{smooth} = \sum_{t=2}^{N}||\hat{\vec{\beta}}_{t}-\hat{\vec{\beta}}_{t-1}||_2^{2} + \sum_{t=2}^{N}||\hat{\vec{\theta}}_{t}-\hat{\vec{\theta}}_{t-1}||_2^{2}, \\
\end{aligned}
\end{equation}
where \(\boldsymbol{\beta}=\{\vec{\beta}_1, \vec{\beta}_2, \ldots, \vec{\beta}_N\}\) and \(\boldsymbol{\theta}=\{\vec{\theta}_1, \vec{\theta}_2, \ldots, \vec{\theta}_N\}\) represent the shape and pose parameters for an N-frame sequence, respectively. 
\(\mathcal{L}_{\text{varen}}\) is a L2 loss on VAREN parameters, \(\mathcal{L}_{\text{2D}}\) and \(\mathcal{L}_{\text{3D}}\) are L1 losses on 2D and 3D keypoints. 
The terms \(\lambda_{\text{varen}}\), \(\lambda_{\text{smooth}}\), \(\lambda_{\text{2D}}\) and \(\lambda_{\text{3D}}\) are hyper-parameters used to balance the influence of each component in the total loss. During training, N is fixed to 16; for inference on arbitrarily long videos, we use a sliding window approach.

\noindent\textbf{Post-Optimization.}
While the spatio-temporal transformer ensures smooth and plausible motion, the predicted mesh may not align perfectly with the image evidence. 
To address this limitation, the second stage of AniMoFormer is a Post-Optimization process. 
We use a differentiable renderer~\citep{ravi2020accelerating} to project the 3D mesh, yielding rendered masks and 2D keypoints. They are compared against pseudo ground truth (GT) 2D keypoints and masks extracted from the input images by ViTPose++~\citep{xu2023vitpose++} and Samurai~\citep{yang2024samurai}, respectively. 
The optimization term is: 
\begin{equation}
\begin{aligned}
    \label{total_loss}
    \mathcal{L} =& \lambda_{\text{2D}} \mathcal{L}_{\text{2D}} + \lambda_{\text{smooth}} \mathcal{L}_{\text{smooth}} + 
    \lambda_{\text{reg}} \mathcal{L}_{\text{reg}} + 
    \lambda_{\text{mask}} \mathcal{L}_{\text{mask}}.
\end{aligned}
\end{equation}
The \(\mathcal{L}_{\text{2D}}\) and \(\mathcal{L}_\text{smooth}\) are defined as in Eq.~\eqref{animoformer_loss}. 
\(\mathcal{L}_{\text{reg}}\) is an L2 regularization term that penalizes unrealistic poses, and \(\mathcal{L}_{\text{mask}}\) is an L1 loss between the rendered masks and the pseudo GT masks.
Both the spatio-temporal transformer and the Post-Optimization are necessary to ensure alignment between the meshes and 2D images according to our ablation studies. More details are in supplementary material. 

\subsection{Appearance Reconstruction with EquineGS}
\label{sec:avatar_reconstuction}
This section outlines \textbf{EquineGS}, a feed-forward approach to reconstructing horse appearance from a single image using 3DGS (illustrated in Fig.~\ref{fig:pipeline}(b)). We first introduce the core components of EquineGS in Sec.~\ref{sec:sec:equinegs} and then describe \textbf{VarenTex} in Sec.~\ref{sec:varentex}, a new multi-view synthetic dataset required to train this appearance network.

\subsubsection{EquineGS}

\label{sec:sec:equinegs}
\noindent\textbf{Canonical Point Cloud Initialization.}
We represent the animal's canonical shape as a 3D point cloud to initialize the positions of the 3D Gaussians. This point cloud is derived from the VAREN template mesh. However, the 13,873 vertices of the base mesh are insufficient to capture fine details. Therefore, we subdivide the mesh by adding a vertex at the midpoint of each edge and splitting each face into four. This upsampling yields \(N_G=55,486\) vertices, which serve as the initial Gaussian positions for EquineGS.

\noindent\textbf{Dual-Stream Feature Extraction.}
Our process begins by extracting features from two parallel streams: one for the \(I \in \mathbb{R}^{448 \times 448}\) input image and one for the initial 3D point cloud. We use a pre-trained DINOv3~\citep{simeoni2025dinov3} backbone (ViT-Large version) to extract multi-scale feature maps that capture both local and global information. These feature maps are then fused via 1x1 convolutions to produce a final image feature map \(\mathbf{F}_{\text{I}} \in \mathbb{R}^{784\times1024}\). In parallel, we process the 3D point coordinates. Following Point Transformer~\citep{zhao2021point}, we apply positional encoding to the points and pass them through an MLP to obtain point features \(\mathbf{F}_{\text{P}} \in \mathbb{R}^{N_G\times1024}\). 

\noindent\textbf{Dual-Stream Transformer Gaussian Decoder.}  
To effectively fuse the multi-modal image feature \(\mathbf{F}_{\text{I}}\) and point feature \(\mathbf{F}_{\text{P}}\) and predict attributes for the Gaussian points, we propose a novel dual-stream transformer gaussian (DSTG) decoder. The DSTG structure (Fig.~\ref{fig:pipeline}(c)) is a modification of the MMDiT block from Qwen-Image~\citep{wu2025qwen}. DSTG operates in three stages:  
\textit{1) Global Context Extraction:} We compute a global context vector \(\mathbf{F}_{\text{context}} \in \mathbb{R}^{1 \times 1024}\) from the image feature map \(\mathbf{F}_{\text{I}}\) for attention modulation via average pooling and an MLP:
\begin{equation}
\begin{aligned}
    \label{equ:global_context}
    \mathbf{F_{\text{context}}} = \text{MLP(AvgPool}(\mathbf{F}_{\text{I}})). 
\end{aligned}
\end{equation}
\textit{2) Feature Fusion:} We simultaneously feed the image features \(\mathbf{F}_{\text{I}}\), point features \(\mathbf{F}_{\text{P}}\), and global context \(\mathbf{F}_{\text{context}}\) into the DSTG decoder. The decoder's core function uses the image information to guide the point features toward an appearance-aligned representation \(\mathbf{F_{\text{DSTG}}}\). 
\textit{3) Attribute Prediction:} Finally, the fused features \(\mathbf{F}_{\text{DSTG}} \in \mathbb{R}^{N_G \times 1024}\) are passed through an MLP to predict the attributes for all \(N_G\) Gaussian points:
\begin{equation}
\begin{aligned}
    \label{equ:gau_attr_pred}
    \{\Delta\boldsymbol{\mu}_i, \mathbf{r}_i, \mathbf{s}_i, \mathbf{c}_i, o_i\}_{i=1}^{N_G} = \text{MLP}(\mathbf{F}_{\text{DSTG}}),
\end{aligned}
\end{equation}
where for each Gaussian \(i\): \(\Delta\boldsymbol{\mu}_i \in \mathbb{R}^{3}\) is the positional offset, \(\mathbf{r}_i \in \mathbb{R}^{4}\) is the rotation, \(\mathbf{s}_i \in \mathbb{R}^{3}\) are the per-axis scale factors, \(\mathbf{c}_i \in \mathbb{R}^{3}\) is the color, and \(o_i \in \mathbb{R}\) is the opacity.

\noindent\textbf{Optimization and Regularization.}
After predicting the canonical Gaussian attributes, we use the shape parameter \(\vec{\beta}\) and pose parameter \(\vec{\theta}\) to deform the Gaussian points into per-frame pose space via the LBS process given by VAREN. 
These points are then rendered from the camera viewpoint (estimated by AniMoFormer) using 3DGS~\citep{kerbl3Dgaussians} to obtain a rendered image \(\hat{I}\) and silhouette mask \(\hat{M}\). The network is trained by minimizing a composite loss function:
\begin{equation}
\begin{aligned}
    \label{equ:loss}
    \mathcal{L} = \lambda_{\text{image}} \mathcal{L}_{\text{image}} + \lambda_{\text{mask}} \mathcal{L}_{\text{mask}} + \lambda_{\text{reg}} \mathcal{L}_{\text{reg}}.
\end{aligned}
\end{equation}

To ensure the visual similarity, we use a combination of the L1 loss and the LPIPS perceptual loss~\cite{zhang2018unreasonable} to capture both pixel-level accuracy and higher-level feature similarity: $\mathcal{L}_{\text{image}} = ||\hat{I}-I||_{1} + \text{LPIPS}(\hat{I}, I)$. 
Additionally, $\mathcal{L}_{\text{mask}}=||\hat{M} - M||_{1}$ enforces silhouette accuracy against the GT mask $M$. 
The terms \(\lambda_{\text{image}}\), \(\lambda_{\text{mask}}\), and \(\lambda_{\text{reg}}\) are hyper-parameters to balance the loss components.

\subsubsection{VarenTex Dataset}
\label{sec:varentex}
While VarenPoser dataset is effective for motion training, its generated texture quality is insufficient for our goal of high-fidelity avatar reconstruction. Moreover, VarenPoser is a video dataset, while training feed-forward avatar network requires multi-view images instead of monocular video. Therefore, to obtain photorealistic multi-view training data for EquineGS, we create a new synthetic dataset called \textbf{VarenTex} using a multi-view diffusion model.

\begin{figure}[ht]
    \centering
    \includegraphics[width=0.8\linewidth]{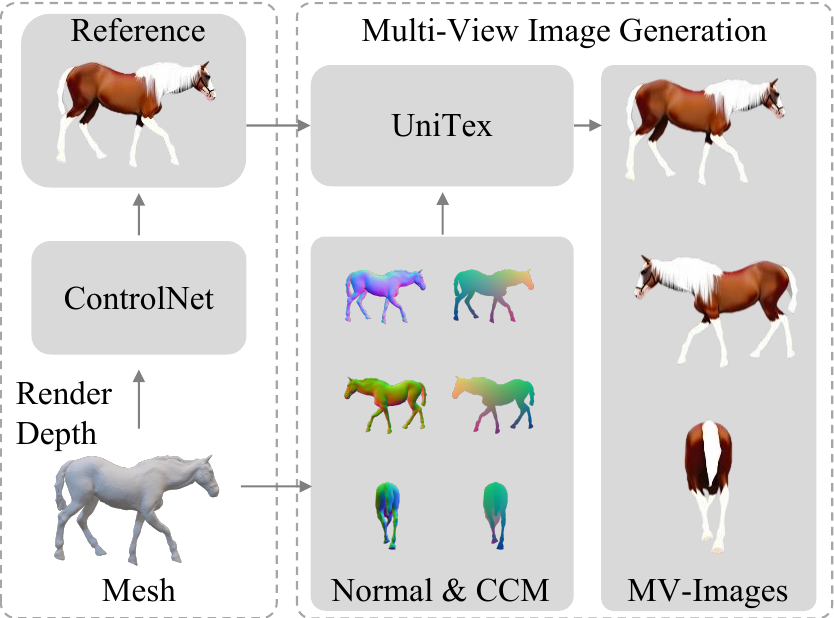}
    \caption{\textbf{VarenTex Generation pipeline.} Normal and canonical coordinate maps (CCM) rendered from VarenPoser meshes , alongside a ControlNet-generated reference image , are fed into UniTex to synthesize multi-view training images.} 
    \label{fig:varentex}
\end{figure}

The construction process (Fig.~\ref{fig:varentex}) employs UniTex~\citep{liang2025unitex}, a model that generates multi-view highly consistent images with the reference image by conditioning on dense geometric information.
First, we acquire 3D meshes from VarenPoser.
Second, from each mesh, we render the geometric conditions required by UniTex: a normal map and a canonical coordinate map (CCM), where each pixel encodes a 3D surface coordinate.
Third, to provide an appearance guide, following~\citep{lyu2025animer, niewiadomski2025ICCV}, we also use ControlNet~\citep{Qwen-Image-ControlNet-Union} to generate a single realistic reference image for appearance guidance.
Finally, these three inputs—the normal map, the coordinate map, and the image—are fed into UniTex to generate the final set of multi-view training images. VarenTex contains 150K multi-view images with a resolution of \(512 \times 512\). 

%% file: sec/05_experiment.tex
\section{Experiment}
\begin{figure*}[ht]
    \centering
    \includegraphics[width=\linewidth]{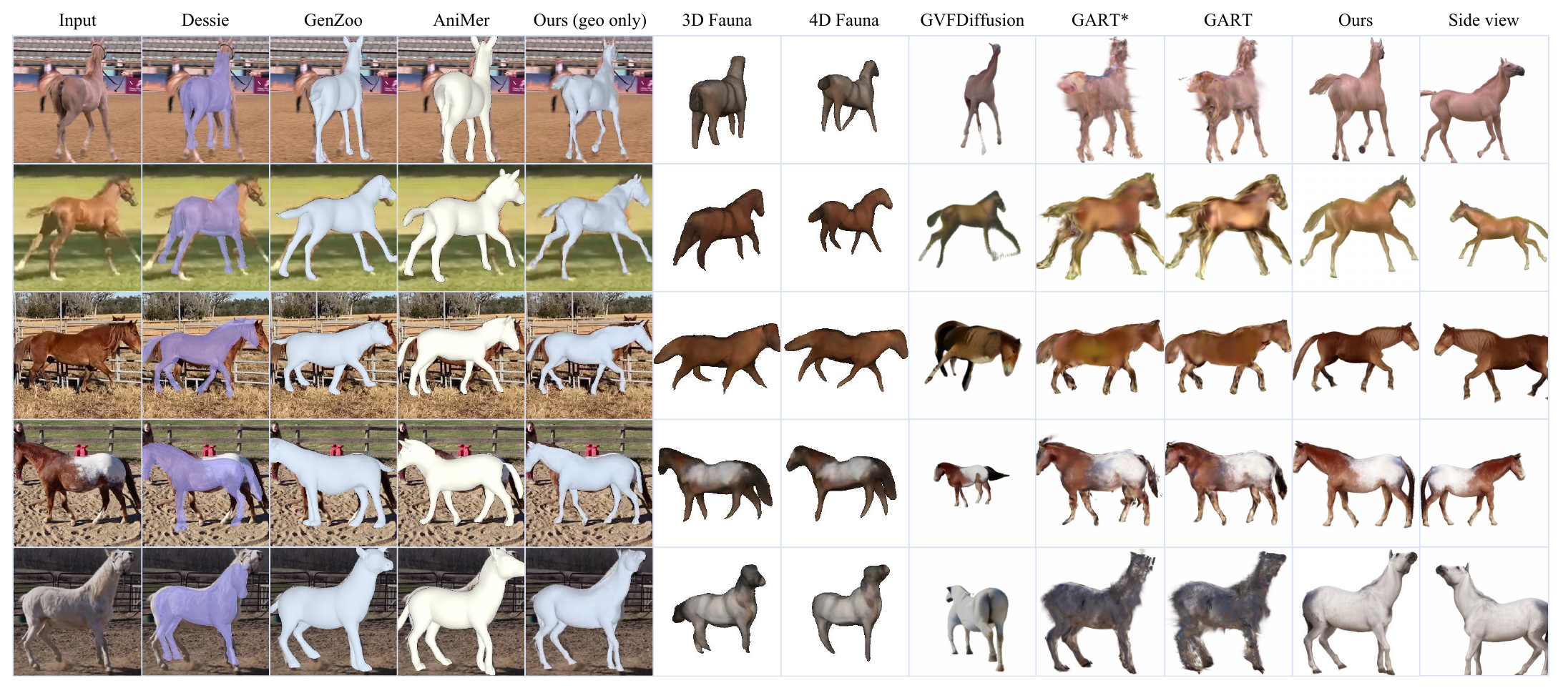}
    \caption{\textbf{Qualitative comparison with the SOTA methods on the AiM dataset.} GART\(^{*}\): Few-shot GART. 
    ``Input'' here is the middle frame of each test video clip. Note that the input image for EquineGS is the first image in the video; therefore, the results of ``Ours'' shown in this figure correspond to novel-pose animation.
    }
    \label{fig:mocap_quali_compare}
\end{figure*}


\subsection{Experimental Setup}
\subsubsection{Datasets}
\textbf{Pose and Shape Estimation.} We train the AniMoFormer using our VarenPoser dataset only. For evaluation, we test our method on three datasets. For assessing 2D keypoint performance and motion smoothness on real-world data, we use two existing benchmarks: APT-36K~\citep{yang2022apt}, a video dataset with over 53,000 instances across 2,400 sequences. AiM~\citep{Animal4D}, another large-scale video dataset. We conduct our evaluation specifically on the horse test subsets of both datasets (30 video clips for APT36K, totaling 402 frames, and 10 video clips for AiM, totaling 453 frames). Finally, to evaluate 3D geometry accuracy, we report performance on the test split of our synthetic VarenPoser dataset (3 video clips, totaling 543 frames), which provides the necessary ground-truth 3D mesh data. 

\noindent\textbf{Avatar Reconstruction.} The EquineGS model is trained on VarenTex only. For evaluation, we test two distinct capabilities using sequences from the AiM dataset:
(1) Novel-View/Pose Reconstruction: We assess this on 35 video sequences of horses, specifically selected to encompass a diverse range of camera views and poses. 
(2) Zero-Shot Generalization: To demonstrate our model's ability to generalize to unseen species, we also select 10 video sequences of zebras that encompass a diverse range of camera views. This represents a challenging zero-shot task because our VarenTex dataset, which is built upon the horse-specific VAREN model, does not include any zebra data.

\subsubsection{Evaluation Metrics}
\textbf{Pose and Shape Estimation.} We evaluate the accuracy of 2D keypoints using the Percentage of Correct Keypoints (PCK). To assess motion smoothness, we compute the acceleration error (Accel). Finally, we measure mesh similarity between the predicted and ground-truth meshes using the Chamfer Distance (CD) after Procrustes alignment.

\noindent\textbf{Avatar Reconstruction.} We evaluate the quality of the novel-view and novel-pose rendered frames using three standard image-comparison metrics following GART~\citep{lei2024gart}: the Peak Signal-to-Noise Ratio (PSNR), the Structural Similarity Index Measure (SSIM), and the Learned Perceptual Image Patch Similarity (LPIPS). 

\subsubsection{Baselines}
\textbf{Pose and Shape Estimation.} 
We benchmark AniMoFormer against five state-of-the-art (SOTA) methods. We select three strong model-based techniques: Dessie~\citep{li2024dessie}, which leverages the specialized hSMAL model~\citep{li2021hsmal} for horses; AniMer~\citep{lyu2025animer}, a transformer-based method built on the SMAL model~\citep{Zuffi:CVPR:2017}; and GenZoo~\citep{niewiadomski2025ICCV}, which employs the SMAL+ variant~\citep{zuffi_eccv2024_awol}. To ensure a comprehensive comparison against different architectural philosophies, we also include 3D-Fauna~\citep{li2024learning} and its extension 4D-Fauna~\citep{Animal4D} as representative SOTA model-free reconstruction methods.

\noindent\textbf{Avatar Reconstruction.} For the evaluation of high-fidelity appearance reconstruction, we compare EquineGS against 3D-Fauna~\citep{li2024learning}, 4D-Fauna~\citep{Animal4D}, GVFDiffusion~\citep{zhang2025gaussian} (a 4D reconstruction method using a pretrained 3D generation model), and GART~\citep{lei2024gart}. Note that, for each video, there is a number of frames used only for testing following GART. Our EquineGS takes only the first training frame to generate appearance; GART uses all the training frames to create avatar; other methods perform frame-by-frame 4D reconstruction on whole video including test frames. 
We follow the fair comparison protocol from~\citep{cho2025dogrecon} to report results against two GART variants: (1) GART, the fully-optimized version following the original paper setting, and (2) GART\(^{*}\), a few-shot GART which is optimized using only 4 training frames for each video. To be fair, the base template of GART is changed to VAREN, and the input pose and shape sequences of GART are the output of AniMoFormer used by 4DEquine. 

\subsubsection{Training Details of 4DEquine}
Due to space limit, training details of 4DEquine are provided in the supplementary materials.

\subsection{Qualitative Evaluation of 4DEquine}
As depicted in Fig.~\ref{fig:mocap_quali_compare} (left), the geometry-only output of 4DEquine (the fifth column) significantly outperforms all baselines in terms of pose and shape recovery, particularly for the leg alignment accuracy. This improved alignment stems from our AniMoFormer's ability to regress smooth and accurate VAREN parameters, which, combined with the VAREN model's inherent detail, produces a more realistic geometric fit than other baselines.

The full 4D reconstruction with appearance in Fig.~\ref{fig:mocap_quali_compare} (right) highlights the power of our disentangled framework. Template-free methods (3D/4D-Fauna) lack a strong geometric prior, resulting in distorted shapes. Optimization-based (GART) and generative (GVFDiffusion) methods also exhibit systematic failure modes on real-world inputs. Specifically, GART often produces artifacts and texture floaters when observations are incomplete, while GVFDiffusion produces inconsistent orientations or global pose ambiguities. In contrast, 4DEquine recovers stable geometry from limited viewpoints and synthesizes full body appearance that better matches the input image. 

Crucially, the high-fidelity 4DEquine results in Fig.~\ref{fig:mocap_quali_compare} are all novel-pose renderings where the appearance is generated by EquineGS using only the first frame of the video as input. This demonstrates a key advantage over methods like GART, which require optimization over many frames. Our feed-forward model generates a complete, animatable avatar from a single view, proving its robustness to the sparse viewpoint coverage typical of real-world recordings. 

\begin{figure}[ht]
    \centering
    \includegraphics[width=0.75\linewidth]{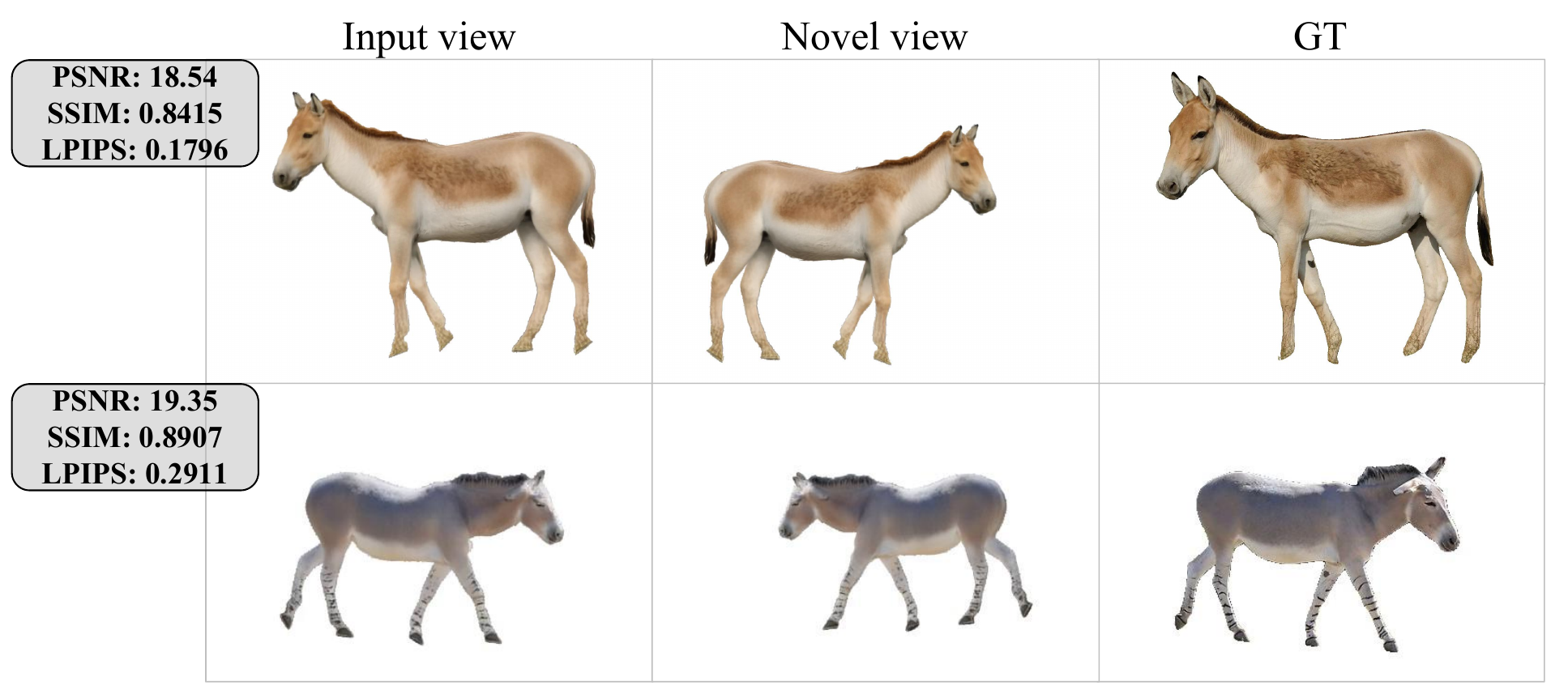}
    \caption{\textbf{Zero-shot generalization of 4DEquine on Internet images of unseen species donkey. ``GT'' is also the input image. }}
    \label{fig:zero_shot}
\end{figure}

Despite 4DEquine being trained exclusively on horses using the VAREN model, it produces plausible 4D reconstructions for unseen equine species like the Equus hemionus and Equus africanus (both are donkeys), as shown in Fig.~\ref{fig:zero_shot}. Note that input images of Fig.~\ref{fig:zero_shot} are from the Internet with backgrounds removed. To enable reconstruction on single image instead of video clip, the input image is tiled with itself to form a N-frame stack input. As a result, 
4DEquine successfully captures the pose and appearance from the single input image, demonstrating that 4DEquine is not merely memorizing the trained horse textures but has learned a robust image feature extraction. 

\subsection{Quantitative Analysis}
\begin{table*}[ht]
\caption{\textbf{Quantitative comparisons on the APT36K dataset, AiM dataset and VarenPoser dataset.}}
\label{tab:mocap_experiment}
\centering
\resizebox{0.85\textwidth}{!}{
\begin{tabular}{cccccccccc}
\toprule
\multirow{2}{*}{Method} & \multicolumn{3}{c}{APT36K} & \multicolumn{3}{c}{AiM} & \multicolumn{3}{c}{VarenPoser} \\ 
\cmidrule(lr){2-4} \cmidrule(lr){5-7} \cmidrule(lr){8-10}  
& PCK@0.05\(\uparrow\) & PCK@0.1\(\uparrow\) & Accel\(\downarrow\) & PCK@0.05\(\uparrow\)  & PCK@0.1\(\uparrow\)  & Accel\(\downarrow\) & PCK@0.05\(\uparrow\)   & Accel\(\downarrow\)  & CD\(\downarrow\)  \\ \hline
3D-Fauna   & 20.1 & 51.4 & 189.3 & 33.3	& 71.8 & 42.3 & 18.9 & 13.8	& 43.0  \\
4D-Fauna   & 25.5 & 53.5 & 177.7 & 46.5 & 74.8 & 32.7 & 28.7 & 11.2 & 38.5 \\
Dessie     & 22.0 & 53.1 & 353.1 & 40.3 & 75.9 & 85.8 & 35.6 & 28.0 & 10.0  \\
GenZoo     & 27.9 & 60.0 & 190.7 & 42.1	& 80.6 & 43.1 & 45.6 & 11.4	& 22.5  \\
AniMer    & 44.5 & 76.6 & 130.5 & 55.5 & 87.7 & 26.2 & 62.2 & 8.4  & 15.2  \\ \midrule
AniMoFormer & \textbf{61.8} & \textbf{83.9} & \textbf{128.6} & \textbf{84.2} & \textbf{95.3} & \textbf{21.8} & \textbf{90.5} & \textbf{2.6}  & \textbf{3.4}   \\ \bottomrule
\end{tabular}
}
\end{table*}

We first evaluate the geometry reconstruction ability of 4DEquine, i.e., AniMoFormer. 
As shown in Tab.~\ref{tab:mocap_experiment}, AniMoFormer outperforms all baselines by a large margin, achieving the highest PCK for pose estimation, the lowest Accel error for motion smoothness, and the lowest CD. 

Further, we compare the appearance quality of 4DEquine with previous 4D reconstruction methods on AiM under the protocol stated before, as is presented in Tab.~\ref{tab:avatar_comp}. On the horse subset, 4DEquine demonstrates superior perceptual and structural quality over model-free reconstruction methods 3D-Fauna, 4D-Fauna and GVFDiffusion significantly due to our better geometric details. We note that the fully-optimized GART model achieves a slightly higher PSNR. This is an expected trade-off: GART's per-video optimization can overfit to each input video and achieve high pixel-level fidelity. However, our method's superior SSIM and LPIPS scores demonstrate that 4DEquine are perceptually and structurally more accurate. 
In addition, our single-image model soundly outperforms the more comparable Few-shot GART across all metrics. On the zebra subset, the advantages of 4DEquine are most evident in the challenging zero-shot generalization task (our VarenTex dataset used for training only contains horse). Our model outperforms all competing methods, including the fully-optimized GART, across all three metrics. 

Moreover, 4DEquine is orders of magnitude more efficient than GART, requiring only 11 seconds per-frame for reconstructing a video. This time is measured on a single NVIDIA A100 GPU, and considers both geometry and appearance reconstruction time. Note that, GART iterates for a fixed 10k steps, resulting in a fixed 15min time consumption independent of frame number used for avatar training.

\begin{table}[ht]
\caption{\textbf{Novel view/pose quantitative comparisons with SOTA methods on the horse and zebra.} GART\(^{*}\): Few-shot GART. \textbf{Bold} and \underline{underlined} numbers indicate the best performance and the second best performance, respectively.}
\label{tab:avatar_comp}
\centering
\resizebox{\columnwidth}{!}{
\begin{tabular}{ccccccccc}
\toprule
\multirow{2}{*}{Method} & \multicolumn{3}{c}{Horse} & \multicolumn{3}{c}{Zebra} \\ \cmidrule(lr){2-4} \cmidrule(lr){5-7}
& PSNR\(\uparrow\) & SSIM\(\uparrow\) & LPIPS\(\downarrow\)  & PSNR\(\uparrow\)  & SSIM\(\uparrow\)   & LPIPS\(\downarrow\)  \\ \hline
3D-Fauna & 12.20          & 0.7205 & 0.2782 & 12.33 & 0.6827 & 0.3318  \\
4D-Fauna & 13.41		  & 0.7550 & 0.2467 & 13.39 & 0.7157 & 0.3055  \\
GVFDiffusion & 12.68 & \underline{0.8189} & 0.2493 & 12.26 & \underline{0.7749} & 0.2897 \\
GART\(^{*}\) & 15.42 & 0.7550 & 0.2452 & 14.31 & 0.6485 & 0.2973  \\
GART     & \textbf{16.19} & 0.7819 & \underline{0.2308} & \underline{15.21} & 0.6752 & \underline{0.2287} \\ \midrule
4DEquine & \underline{15.66} & \textbf{0.8364} & \textbf{0.1720} & \textbf{15.54} & \textbf{0.7828} & \textbf{0.2000} \\ \bottomrule
\end{tabular}
}
\end{table}

\subsection{Ablation Study}
\begin{table}[ht]
\caption{\textbf{Ablation study for AniMoFormer.} 
}
\label{tab:mocap_ablation}
\centering
\resizebox{0.9\columnwidth}{!}{
\begin{tabular}{ccccc}
\toprule
\multirow{2}{*}{Variant} & \multicolumn{2}{c}{APT36K} & \multicolumn{2}{c}{AiM} \\ \cmidrule(lr){2-3} \cmidrule(lr){4-5} 
 & PCK@0.05\(\uparrow\) & Accel\(\downarrow\) & PCK@0.05\(\uparrow\)  & Accel\(\downarrow\) \\ \hline
 w/o PO \& Temporal & 37.1 & 134.7 & 45.1 & 30.6 \\
 w/o PO    & 37.7 & 129.1 & 47.8  & 25.7       \\
 w/o Temporal   & 57.9  & 143.2 & 82.9  & 24.7   \\
 AniMoFormer & \textbf{61.8}  & \textbf{128.6} & \textbf{84.2} & \textbf{21.8}       \\ \bottomrule
\end{tabular}
}
\end{table}

\noindent\textbf{Effectiveness of Post-Optimization.} 
We conduct an ablation study to analyze the quantitative and qualitative effects of our Post-Optimization design, with results shown in Tab.~\ref{tab:mocap_ablation}. 
We compare the full AniMoFormer with the one without Post-Optimization (labeled ``w/o PO''). 
The results show a clear qualitative improvement of PO, as the final reconstructions align more closely with the 2D image evidence. 
Moreover, the results in Tab.~\ref{tab:avatar_ablation} (``w/o PO'') show a significant degradation in visual quality across all rendering metrics without PO. This is intuitive: mesh misalignment would reduce the final 4D reconstruction quality. 

\noindent\textbf{Effectiveness of Temporal Transformer.} 
The effectiveness of our temporal transformer is most evident in its contribution to motion smoothness. This is quantitatively confirmed in Tab.~\ref{tab:mocap_ablation}, where removing the temporal transformer (``w/o Temporal'') results in a higher acceleration error, indicating less plausible motion. 


\begin{table}[ht]
\caption{\textbf{Ablation study of EquineGS components and post-optimization for novel-pose appearance reconstruction on the AiM horse subset.} 
\textbf{Bold} and \underline{underlined} numbers indicate the best performance and the second best performance, respectively.}
\label{tab:avatar_ablation}
\centering
\resizebox{0.65\columnwidth}{!}{
\begin{tabular}{cccc}
\toprule
\multirow{2}{*}{Variant} & \multicolumn{3}{c}{Horse} \\ \cmidrule(lr){2-4} 
& PSNR\(\uparrow\) & SSIM\(\uparrow\) & LPIPS\(\downarrow\)  \\ \hline
w/o PO & 13.84 & 0.8103 & 0.2170  \\
w/o SubDiv & \textbf{15.76} & 0.8237 & 0.1871  \\
w/o DSTG & 15.53 & \underline{0.8353} & \underline{0.1733} \\
4DEquine & \underline{15.66} & \textbf{0.8364} & \textbf{0.1720} \\ \bottomrule
\end{tabular}
}
\end{table}

\noindent\textbf{Effect of EquineGS Model Design.} 
To validate the architectural design of our avatar reconstruction network, we compare our full EquineGS model, which uses the proposed DSTG decoder, against a variant named ``w/o DSTG''. In this variant, we replaced our DSTG-Block with a standard cross-attention block for feature fusion, just like a typical Transformer decoder does. As shown in Tab.~\ref{tab:avatar_ablation}, our full model with the DSTG-Block consistently outperforms the Cross-Attn variant, achieving higher PSNR, SSIM and lower LPIPS scores on the horse subset of AiM. 

\begin{figure}[ht]
    \centering
    \includegraphics[width=0.75\linewidth]{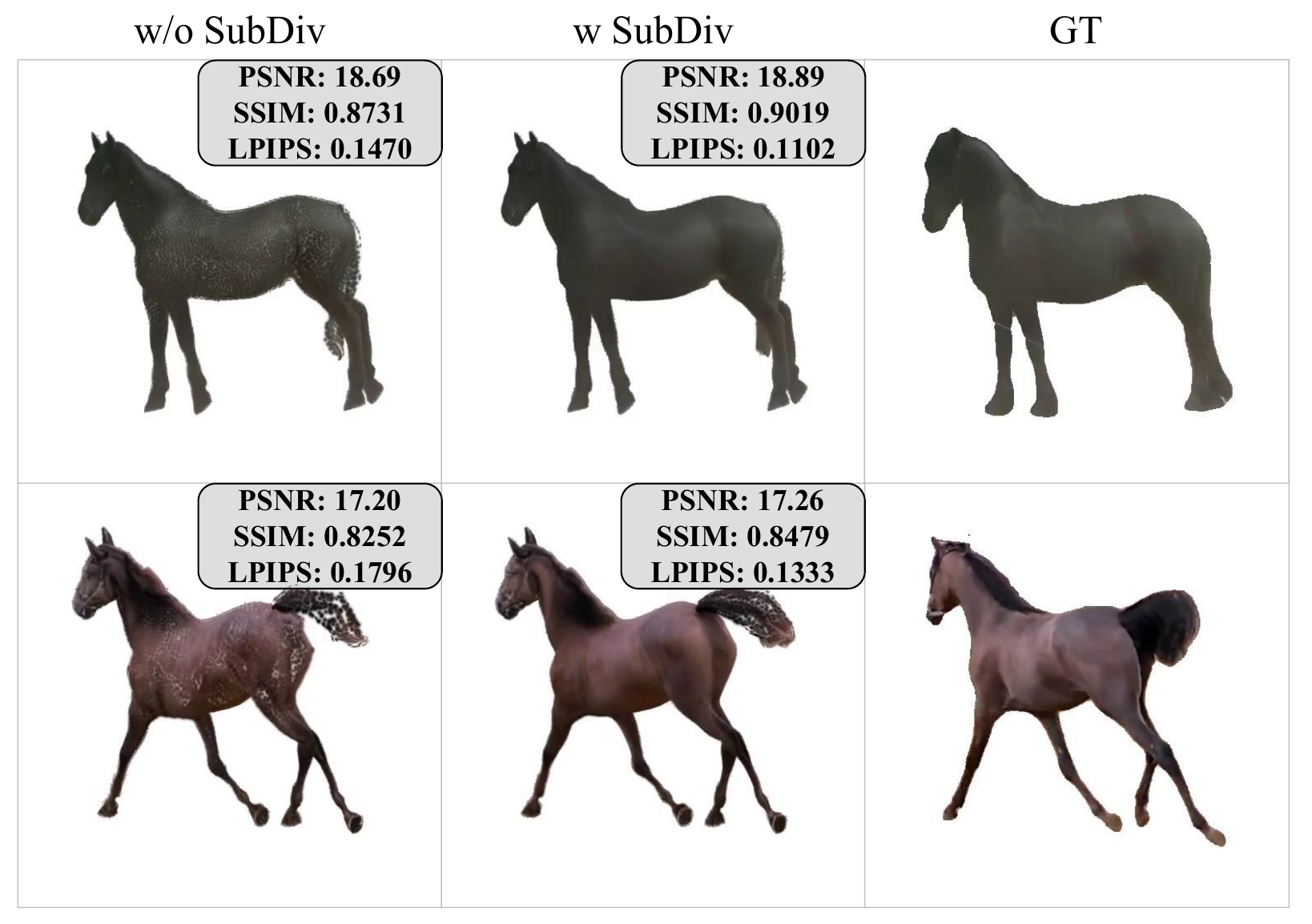}
    \caption{\textbf{Ablation study for Sub-Division.}}
    \label{fig:subdiv_ablation}
\end{figure}

\noindent\textbf{Effect of Different Numbers of Gaussian Points.} To analyze the impact of point cloud density, we compare our full EquineGS model (which uses 55,486 points generated via mesh subdivision) against a ``w/o SubDiv'' variant which uses only the 13,873 vertices of the base VAREN mesh as initial point cloud. As shown in Tab.~\ref{tab:avatar_ablation}, the ``w/o SubDiv'' variant surprisingly achieves a slightly higher PSNR score. However, this metric is misleading. The final visual effect of the low-resolution model is plagued by numerous gaps and holes. As shown in Fig.~\ref{fig:subdiv_ablation}, the sparse 13,873 points are insufficient to form a contiguous surface, resulting in a rendered avatar that looks porous and incomplete.

%% file: sec/06_conclusion.tex
\section{Conclusion}
\noindent\textbf{Summary.} 
We present 4DEquine, an efficient framework for 4D equine reconstruction from monocular video. Our key contribution is disentangling the 4D task into dynamic motion estimation and static appearance reconstruction, bridged by the VAREN model. For motion, AniMoFormer combines a spatio-temporal transformer for smooth regression with a post-optimization stage for 2D alignment. For appearance, EquineGS is a feed-forward network that generates a high-fidelity, animatable 3D Gaussian avatar from a single image, bypassing costly per-video optimization. We support this with two new synthetic datasets, VarenPoser (motion) and VarenTex (appearance). Experiments show 4DEquine achieves state-of-the-art performance and strong zero-shot generalization to other species like donkeys and zebras.

\noindent\textbf{Limitations and Future Work.} 
Our synthetic datasets build upon the VAREN model and inherit its limitations, which does not adequately represent the complex physics and appearance of the tail and mane. Additionally, our approach cannot account for variations in environmental conditions, such as dynamic lighting. 
We plan to address these limitations by enhancing the avatar with physics-based representations for tails and manes and adding relighting module. We also aim to improve EquineGS by fusing information from a few keyframes to capture unique markings efficiently. 

\noindent\textbf{Acknowledgements.} This study was supported by the National Key Research and Development Program of China (2023YFC2415400); the National Natural Science Foundation of China (T2422012, 62125107); the Guangdong Basic and Applied Basic Research (2024B1515020088); the High Level of Special Funds (G030230001, G03034K003); the Guangdong Key Research and Development Program (2025B1111080001); the SUSTech Fang Keng Faculty Award.

%% file: sec/X_suppl.tex
\clearpage
\setcounter{page}{1}
\maketitlesupplementary

\section{More Qualitative Results}
\begin{figure*}[ht]
    \centering
    \includegraphics[width=\linewidth]{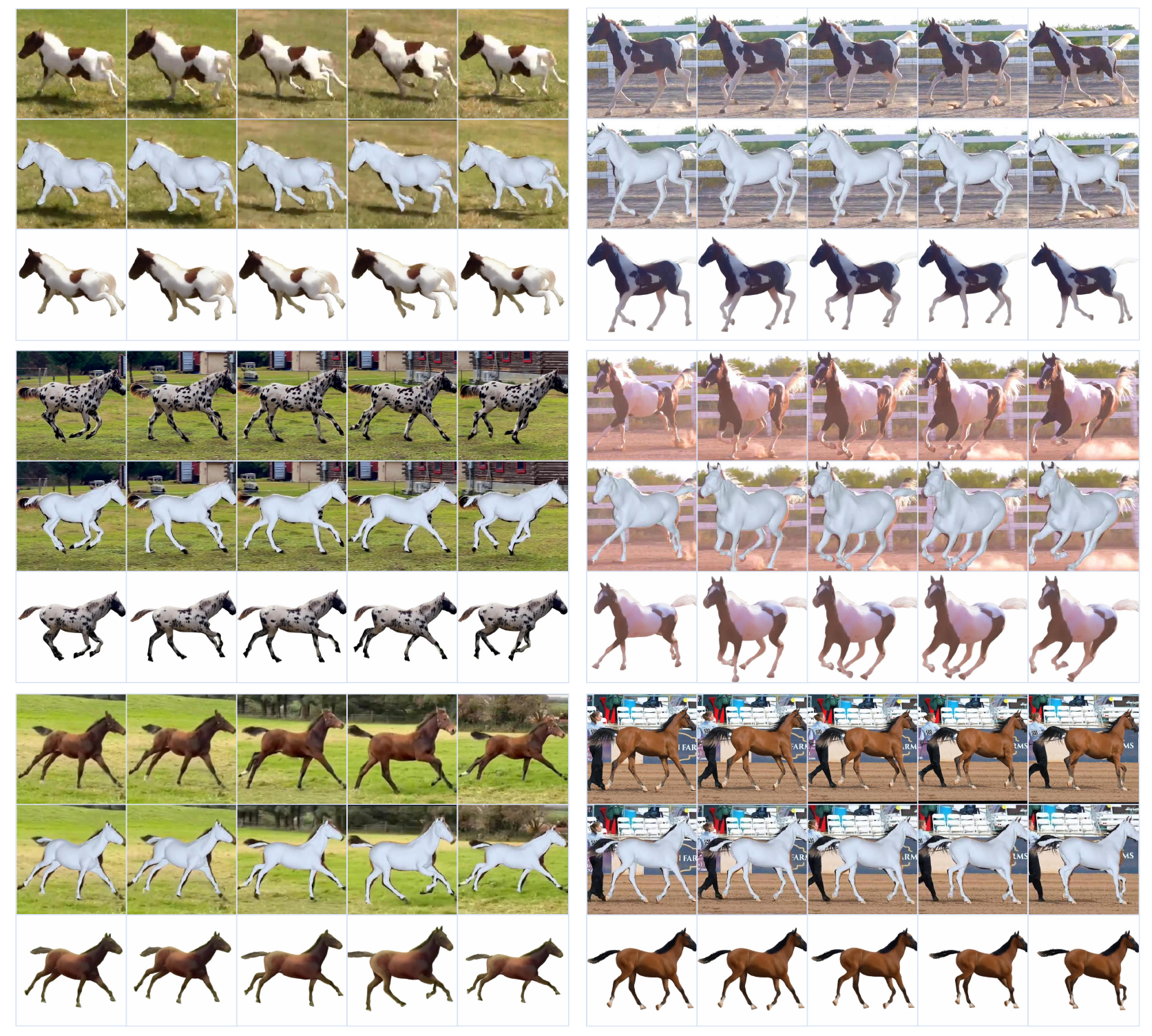}
    \caption{\textbf{More Qualitative Results of 4DEquine.} In each example, the first row displays the reference video frames at various time steps, the second row visualizes the output of AniMoFormer, and the third row presents the final reconstruction from EquineGS.}
    \label{fig:more_quali_results}
\end{figure*}

We present additional qualitative results in Fig.~\ref{fig:more_quali_results}. The video sequences were collected from the Internet. For these visualizations, the input to 4DEquine is the first frame of the video; Therefore, the results shown in this figure are the novel-pose animation.

\section{More Details about 4DEquine}
\subsection{Overview of 4DEquine}
\subsubsection{Training Stage}
Due to the scarcity of high-quality 4D equine data, both components of 4DEquine are trained exclusively on synthetic datasets. Motion: AniMoFormer is trained on VarenPoser, our large-scale synthetic video dataset, to regress the VAREN model's pose and shape parameters. Appearance: EquineGS is trained on VarenTex, a synthetic multi-view image dataset, to predict the attributes of canonical 3D Gaussian primitives.
\subsubsection{Inference Stage} 
During inference, 4DEquine processes a real-world monocular video of arbitrary length. The pipeline operates as follows: (1) Motion Recovery: The video is processed by AniMoFormer using a sliding-window strategy to extract temporally smooth VAREN parameters for each frame. (2) Post-Optimization: Because the feed-forward motion predictions may not perfectly align with the image evidence, we apply a brief post-optimization step using 2D keypoints and masks to ensure pixel-aligned equine geometry. (2) Appearance Generation: Concurrently, EquineGS takes a single representative keyframe (e.g., the first frame of the video) and processes it in a feed-forward manner to output a high-fidelity, canonical 3D Gaussian avatar. (3) 4D Reconstruction: Finally, the canonical Gaussian points are deformed into the per-frame pose space using Linear Blend Skinning (LBS) driven by the optimized VAREN parameters, yielding the final 4D reconstruction.

\subsection{More Details about VarenPoser Dataset}
We collected 500 horse images from the Internet, spanning diverse categories, to serve as appearance guidance for MV-Adapter~\citep{huang2024mvadapter}. To ensure appearance diversity within the VarenPoser dataset, we randomly assigned a generated texture to each of the 1,171 video clips. To maximize this diversity, we enforced a constraint limiting the reuse of any unique texture to a maximum of three clips (\(1171/500=2.342\)). The categorical breakdown and image counts for this 500-image guidance set are detailed in Tab.~\ref{tab:num_images}.

\begin{table}[ht]
\caption{\textbf{The number of different categories of horse images.}}
\label{tab:num_images}
\centering
\resizebox{\columnwidth}{!}{
\begin{tabular}{lclc}
\toprule
Category & Number & Category & Number \\ \hline
Black Shire & 22  & Perlino Akhal Teke & 29 \\
Black Tobiano Saddlebred & 40 & Palomino Quarter Horse & 40 \\
Skewbald Shetland Pony & 15 & White Arabian  & 28 \\
Pinto Paint Horse & 50 & Chestnut Morgan  & 18 \\
Black Friesian & 23 & Buckskin Tennessee Walker & 18 \\
Bay Thoroughbred & 41 & Dapple Gray Andalusian & 48 \\
Gray Percheron & 5 & Grullo Dun Mustang & 14 \\
Leopard Appaloosa & 58 & Silver Dapple Icelandic Horse & 16 \\
Gray Lipizzan & 35 & Total & 500 \\ \bottomrule
\end{tabular}
}
\end{table}

\begin{figure}[ht]
    \centering
    \includegraphics[width=\linewidth]{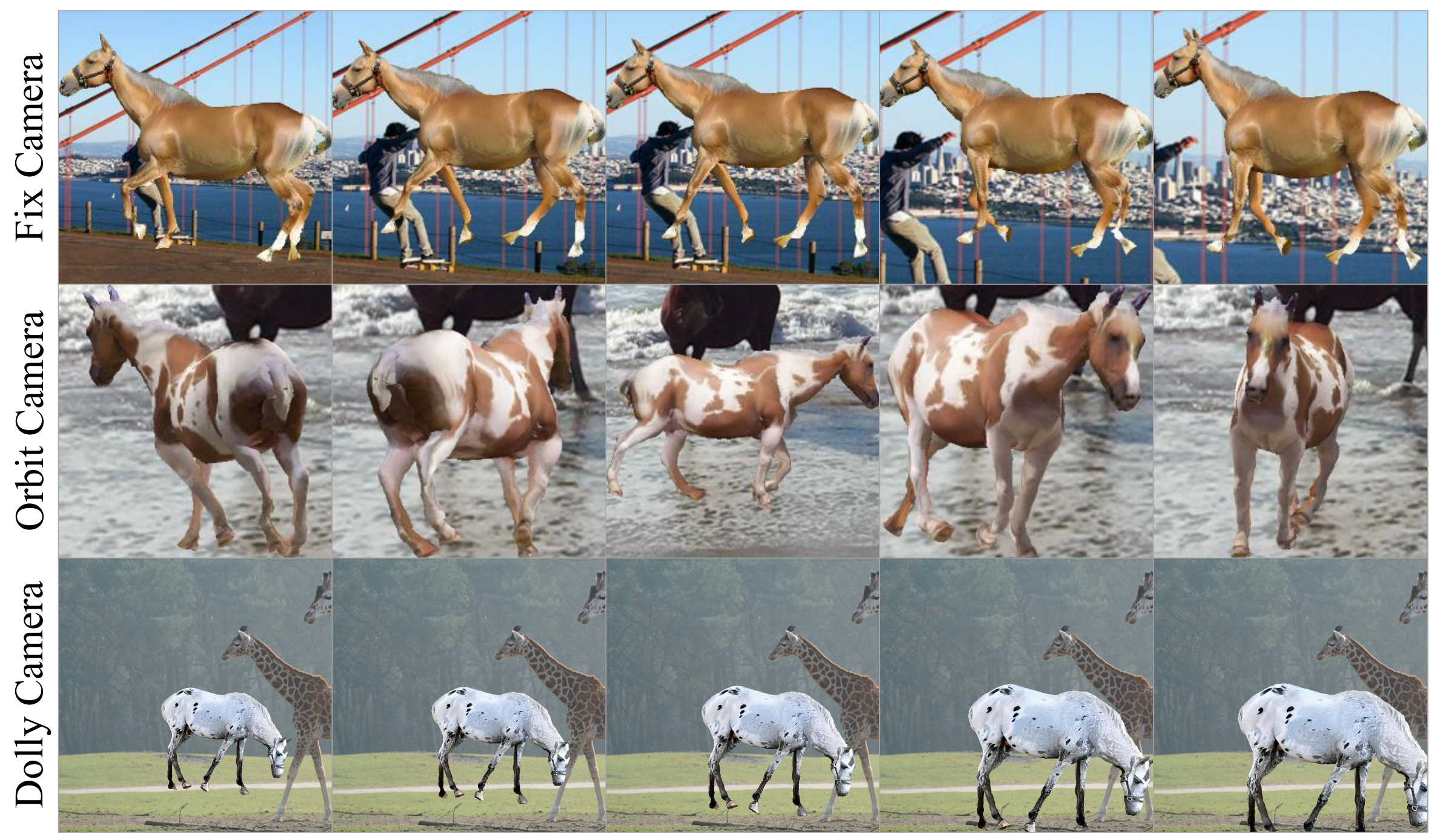}
    \caption{\textbf{Visualization of camera trajectory settings in the VarenPoser dataset.} We illustrate the three camera behaviors used during data generation: Top: Fixed camera; Middle: Orbit camera; Bottom: Dolly camera (The camera moves farther away from or closer to the object). Background images are sampled from the COCO dataset.}
    \label{fig:varenposer2}
\end{figure}

Furthermore, to ensure viewpoint diversity, the initial camera pitch and azimuth angles were randomly sampled within the ranges of \((-15^{\circ}, 15^{\circ})\) and \((-180^{\circ}, 180^{\circ})\), respectively. The camera trajectory settings—fix, orbit, and dolly—were sampled with probabilities of 0.4, 0.3, and 0.3, respectively. Additional visualizations of these camera settings are provided in Fig.~\ref{fig:varenposer2} and background images were sourced from the COCO dataset ~\citep{lin2014microsoft}. 

Consistent with the PFERD~\citep{li2024poses} dataset, VarenPoser maintains a frame rate of 60 frames per second (fps). The video resolution is set to \(512 \times 512\). Each frame is annotated with VAREN parameters, 21 3D keypoints, 21 2D keypoints (including visibility flags), and a segmentation mask.

\subsection{More Details about AniMoFormer}
\subsubsection{Implementation Details}
\textbf{Spatio-Temporal Transformer.}
Our spatial transformer employs a Vision Transformer-Huge (ViT-H)~\citep{dosovitskiy2021imageworth16x16words} backbone, with weights initialized from the pre-trained AniMer~\citep{lyu2025animer}. The temporal transformer is constructed using a standard transformer encoder architecture~\citep{Vaswani2017AttentionIA}. We train the model using the AdamW~\citep{adamw} optimizer with an initial learning rate of \(1\times 10^{-5}\). To enhance robustness to occlusion, we employ a copy-paste data augmentation, randomly overlaying object masks from~\citep{Pont-Tuset_arXiv_2017} with a probability of 0.3. 
Additionally, we simulate varying video frame rates during training by randomly sampling frames at different temporal strides. The model is trained for 100,000 steps on a single NVIDIA RTX 4090 GPU, requiring approximately 10 hours. The final loss components are balanced with the following weights: \(\lambda_{\text{varen}}=0.01\), \(\lambda_{\text{smooth}}=0.001\), \(\lambda_{\text{2D}}=0.05\) and \(\lambda_{\text{3D}}=0.05\). 

\noindent\textbf{Post-Optimization.} 
We employ a two-stage post-optimization process to refine the results, with each stage consisting of 100 iterations. Both stages utilize the AdamW optimizer with a learning rate of \(5\times 10^{-3}\). The two stages are designed to sequentially refine the fit, first by aligning to keypoints and then by fitting the silhouette. In the first stage, we optimize all VAREN parameters and the camera. The loss weights are set to heavily prioritize keypoint accuracy: \(\lambda_{\text{2D}}=10000\), \(\lambda_{\text{smooth}}=100\), \(\lambda_{\text{mask}}=100\), and \(\lambda_{reg}=1000\). In the second stage, we freeze the pose and camera parameters, allowing the optimization to focus on refining the shape. The loss weights are adjusted to prioritize mask alignment: \(\lambda_{\text{2D}}=100\), \(\lambda_{\text{smooth}}=100\), \(\lambda_{\text{mask}}=10000\), \(\lambda_{reg}=300\). 

\subsection{More Details about EquineGS}
Throughout the training process, the DINOv3 backbone remains frozen. For each training instance, we randomly select a single view as input and utilize four additional views for supervision. The model is optimized using the AdamW optimizer with an initial learning rate of \(2\times 10^{-4}\), which includes a linear warmup phase over the first 3,000 steps. We train the model for 100,000 steps using a gradient accumulation of 4 and an image resolution of \(448 \times 448\). The entire training process is distributed across eight NVIDIA RTX 4090 GPUs and requires approximately 3 days to complete. To balance the objective function, the loss hyperparameters are set as follows: \(\lambda_{\text{image}}=1.0\), \(\lambda_{\text{mask}}=0.5\), and \(\lambda_{\text{reg}}=0.1\).

\subsection{Visualization for VarenTex}
\begin{figure}[ht]
    \centering
    \includegraphics[width=\linewidth]{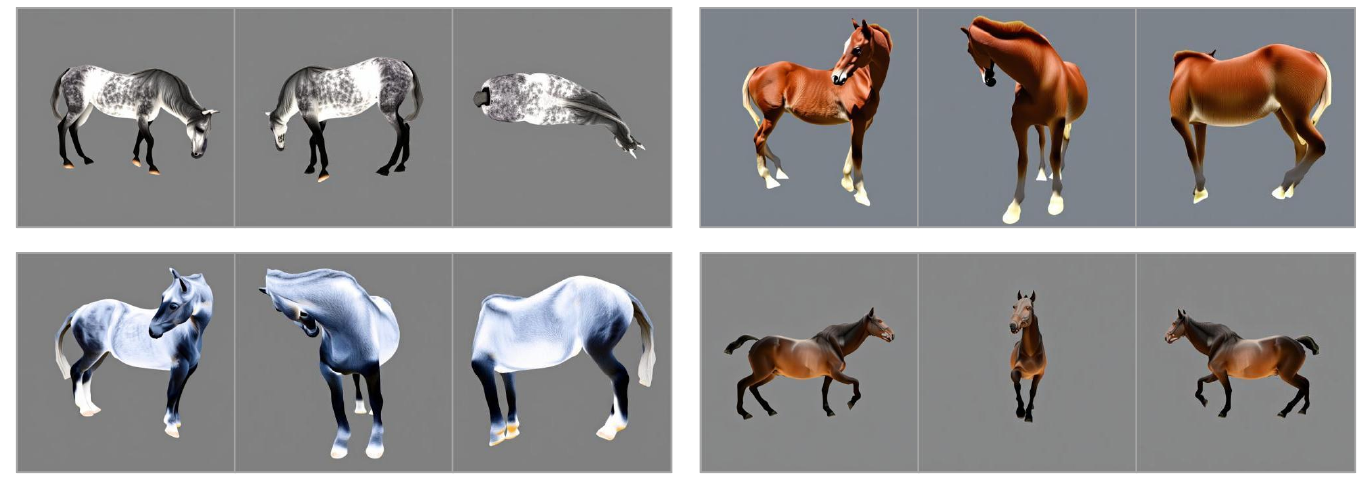}
    \caption{\textbf{VarenTex visual samples.}} 
    \label{fig:varentex_dataset}
\end{figure}
We present representative visual samples from the VarenTex dataset in Fig.~\ref{fig:varentex_dataset}.

\section{More Experiments}
\subsection{Comparison with GART}
We further evaluate pose estimation performance by comparing our method with GART~\cite{lei2024gart} on the AiM dataset, because GART optimizes underlying mesh pose parameters together with its Gaussian appearance using a photometric loss. As detailed in Tab.~\ref{tab:pose_comp_with_gart}, AniMoFormer consistently outperforms GART across all metrics. It is important to note that for fair comparison, GART utilizes the estimation results from AniMoFormer as its initialization. We observe that this optimization process inadvertently degrades pose estimation performance because the photometric loss prioritizes pixel-level color alignment (i.e., rendered image quality) over geometric fidelity. 
This result further proves the validity and importance of our decomposition design. 
\begin{table}[ht]
\caption{\textbf{Quantitative comparisons on the AiM dataset.} GART*: Few-shot GART.}
\label{tab:pose_comp_with_gart}
\centering
\resizebox{\columnwidth}{!}{
\begin{tabular}{ccccccc}
\toprule
 &
  \multicolumn{3}{c}{Horse} &
  \multicolumn{3}{c}{Zebra} \\ \cmidrule(lr){2-4} \cmidrule(lr){5-7} 
\multirow{-2}{*}{Method} &
  PCK@0.05\(\uparrow\) &
  PCK@0.1\(\uparrow\) &
  Accel\(\downarrow\) &
  PCK@0.05\(\uparrow\) &
  PCK@0.1\(\uparrow\) &
  Accel\(\downarrow\) \\ \hline
GART* &
  79.0 &
  97.1 &
  30.7 &
  79.5 &
  95.8 &
  27.8 \\
GART &
  81.8 &
  97.5 &
  29.1 &
  82.5 &
  95.7 &
  27.3 \\
4DEquine &
  \textbf{87.9} &
  \textbf{98.6} &
  \textbf{22.4} &
  \textbf{89.0} &
  \textbf{96.8} &
  \textbf{19.9} \\ \bottomrule
\end{tabular}
}
\end{table}

\subsection{Comparison with AniMer + PO}
To highlight the effectiveness of our model design, we compare AniMoFormer against AniMer~\cite{lyu2025animer} with Post-Optimization. AniMoFormer outperforms ``AniMer + PO'' (Tab.~\ref{tab:comparison_with_AniMerPO}) due to our temporal attention and VAREN's superior expressiveness over SMAL.

\begin{table}[ht]
\caption{\textbf{Compared to AniMer + PO.}}
\label{tab:comparison_with_AniMerPO}
\centering
\resizebox{0.9\columnwidth}{!}{
\begin{tabular}{ccccc}
\toprule
\multirow{2}{*}{Variant} & \multicolumn{2}{c}{APT36K} & \multicolumn{2}{c}{AiM} \\ \cmidrule(lr){2-3} \cmidrule(lr){4-5} 
 & PCK@0.05\(\uparrow\) & Accel\(\downarrow\) & PCK@0.05\(\uparrow\)  & Accel\(\downarrow\) \\ \hline
AniMer + PO & 61.1 & 129.5 & 78.4 & 24.4 \\
AniMoFormer & \textbf{61.8}  & \textbf{128.6} & \textbf{84.2} & \textbf{21.8} \\
\bottomrule
\end{tabular}
}
\end{table}

\subsection{Ablation on the Number of Input Frames}
We perform an ablation for window size (\(N\)) on the AiM dataset. The results demonstrate consistent performance gains as \(N\) increases from 4 to 8 and 16: PCK@0.05 rises from 46.8 to 47.5 and 47.8, while acceleration error decreases from 27.0 to 25.8 and 25.7. However, setting \(N=32\) leads to out-of-memory error. To accommodate videos longer than 16 frames, we employ a sliding-window approach to enable inference over sequences of arbitrary lengths.

\subsection{Qualitative Results of Challenging Pose}
\begin{figure}[ht]
    \centering
    \includegraphics[width=1\linewidth]{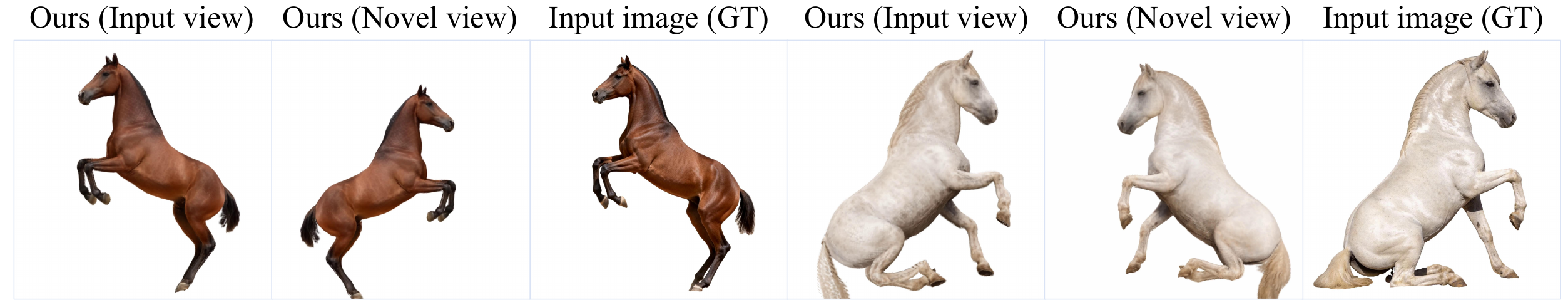}
    \caption{\textbf{Challenging poses. Left: Rearing. Right: Sitting.}}
    \label{fig:challenge_pose}
\end{figure}

We provide more challenging poses in Fig.~\ref{fig:challenge_pose} to demonstrate the robustness of 4DEquine.

\section{Failure Cases and Discussion}
\begin{figure}[ht]
    \centering
    \includegraphics[width=\linewidth]{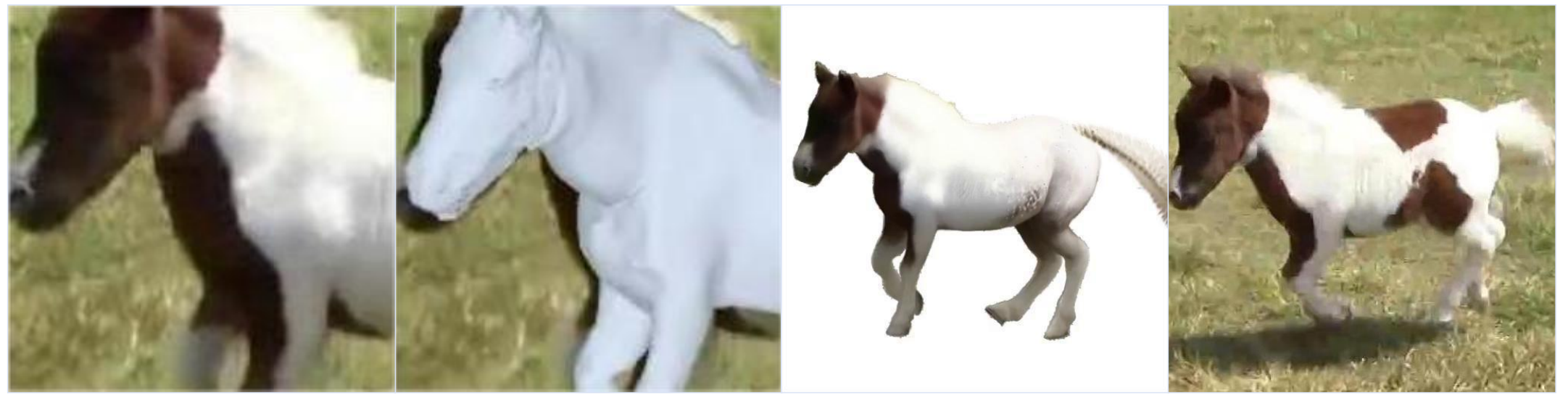}
    \caption{\textbf{Analysis of failure cases.} We illustrate a scenario where severe occlusion impacts performance. The first image: The truncated/occluded input image. The second image: The output of AniMoFormer. The third image: The output of EquineGS, showing appearance degradation at unseen areas. The fourth image: Reference frame from a different time step showing the horse's true appearance.}
    \label{fig:failure_case}
\end{figure}

We illustrate a failure case in Fig.~\ref{fig:failure_case}. Although AniMoFormer is capable of reconstructing geometry when input images suffer from severe truncation or occlusion (as it incorporates data augmentation for occlusion and truncation during training), EquineGS struggles to reconstruct a consistent appearance. Therefore, when selecting keyframes as input for EquineGS, it is necessary to ensure that the horse in the image is not significantly occluded or truncated. In future work, we plan to address these challenges by developing a method for the efficient fusion of multiple keyframes, allowing the model to aggregate appearance information from unoccluded views.

%% file: main.bbl
\begin{thebibliography}{61}
\providecommand{\natexlab}[1]{#1}
\providecommand{\url}[1]{\texttt{#1}}
\expandafter\ifx\csname urlstyle\endcsname\relax
  \providecommand{\doi}[1]{doi: #1}\else
  \providecommand{\doi}{doi: \begingroup \urlstyle{rm}\Url}\fi

\bibitem[An et~al.(2025)An, Lyu, Lin, Cheng, Liu, and Tang]{lyu2025animerunifiedposeshape}
Liang An, Jin Lyu, Li Lin, Pujin Cheng, Yebin Liu, and Xiaoying Tang.
\newblock Animer+: Unified pose and shape estimation across mammalia and aves via family-aware transformer.
\newblock \emph{IEEE Transactions on Pattern Analysis and Machine Intelligence}, pages 1--18, 2025.

\bibitem[Biggs et~al.(2018)Biggs, Roddick, Fitzgibbon, and Cipolla]{biggs2018creatures}
Benjamin Biggs, Thomas Roddick, Andrew Fitzgibbon, and Roberto Cipolla.
\newblock {C}reatures great and {SMAL}: {R}ecovering the shape and motion of animals from video.
\newblock In \emph{ACCV}, 2018.

\bibitem[Cho et~al.(2025)Cho, Kang, Soon, and Joo]{cho2025dogrecon}
Gyeongsu Cho, Changwoo Kang, Donghyeon Soon, and Kyungdon Joo.
\newblock Dogrecon: Canine prior-guided animatable 3d gaussian dog reconstruction from a single image.
\newblock \emph{International Journal of Computer Vision}, 133\penalty0 (9):\penalty0 6332--6346, 2025.

\bibitem[Dosovitskiy et~al.(2021)Dosovitskiy, Beyer, Kolesnikov, Weissenborn, Zhai, Unterthiner, Dehghani, Minderer, Heigold, Gelly, Uszkoreit, and Houlsby]{dosovitskiy2021imageworth16x16words}
Alexey Dosovitskiy, Lucas Beyer, Alexander Kolesnikov, Dirk Weissenborn, Xiaohua Zhai, Thomas Unterthiner, Mostafa Dehghani, Matthias Minderer, Georg Heigold, Sylvain Gelly, Jakob Uszkoreit, and Neil Houlsby.
\newblock An image is worth 16x16 words: Transformers for image recognition at scale, 2021.

\bibitem[Feng et~al.(2025)Feng, Zhang, Wang, Ye, Yu, Black, Darrell, and Kanazawa]{feng2025st4rtrack}
Haiwen Feng, Junyi Zhang, Qianqian Wang, Yufei Ye, Pengcheng Yu, Michael~J Black, Trevor Darrell, and Angjoo Kanazawa.
\newblock St4rtrack: Simultaneous 4d reconstruction and tracking in the world.
\newblock \emph{arXiv preprint arXiv:2504.13152}, 2025.

\bibitem[Huang et~al.(2024)Huang, Guo, Wang, Yi, Ma, Cao, and Sheng]{huang2024mvadapter}
Zehuan Huang, Yuanchen Guo, Haoran Wang, Ran Yi, Lizhuang Ma, Yan-Pei Cao, and Lu Sheng.
\newblock Mv-adapter: Multi-view consistent image generation made easy.
\newblock \emph{arXiv preprint arXiv:2412.03632}, 2024.

\bibitem[InstantX(2025)]{Qwen-Image-ControlNet-Union}
InstantX.
\newblock Qwen-image-controlnet-union.
\newblock \url{https://huggingface.co/InstantX/Qwen-Image-ControlNet-Union}, 2025.

\bibitem[Jakab et~al.(2024)Jakab, Li, Wu, Rupprecht, and Vedaldi]{jakab2024farm3d}
Tomas Jakab, Ruining Li, Shangzhe Wu, Christian Rupprecht, and Andrea Vedaldi.
\newblock Farm3d: Learning articulated 3d animals by distilling 2d diffusion.
\newblock In \emph{2024 International Conference on 3D Vision (3DV)}, pages 852--861. IEEE Computer Society, 2024.

\bibitem[Kanazawa et~al.(2018)Kanazawa, Tulsiani, Efros, and Malik]{kanazawa2018learning}
Angjoo Kanazawa, Shubham Tulsiani, Alexei~A Efros, and Jitendra Malik.
\newblock Learning category-specific mesh reconstruction from image collections.
\newblock In \emph{Proceedings of the European conference on computer vision (ECCV)}, pages 371--386, 2018.

\bibitem[Karaev et~al.(2024)Karaev, Rocco, Graham, Neverova, Vedaldi, and Rupprecht]{karaev2024cotracker}
Nikita Karaev, Ignacio Rocco, Benjamin Graham, Natalia Neverova, Andrea Vedaldi, and Christian Rupprecht.
\newblock Cotracker: It is better to track together.
\newblock In \emph{European conference on computer vision}, pages 18--35. Springer, 2024.

\bibitem[Kaye et~al.(2025)Kaye, Jakab, Wu, Rupprecht, and Vedaldi]{kaye2025dualpm}
Ben Kaye, Tomas Jakab, Shangzhe Wu, Christian Rupprecht, and Andrea Vedaldi.
\newblock {DualPM}: Dual posed-canonical point maps for {3D} shape and pose reconstruction.
\newblock In \emph{CVPR}, 2025.

\bibitem[Kerbl et~al.(2023)Kerbl, Kopanas, Leimk{\"u}hler, and Drettakis]{kerbl3Dgaussians}
Bernhard Kerbl, Georgios Kopanas, Thomas Leimk{\"u}hler, and George Drettakis.
\newblock 3d gaussian splatting for real-time radiance field rendering.
\newblock \emph{ACM Transactions on Graphics}, 42\penalty0 (4), 2023.

\bibitem[Lei et~al.(2024)Lei, Wang, Pavlakos, Liu, and Daniilidis]{lei2024gart}
Jiahui Lei, Yufu Wang, Georgios Pavlakos, Lingjie Liu, and Kostas Daniilidis.
\newblock Gart: Gaussian articulated template models.
\newblock In \emph{Proceedings of the IEEE/CVF conference on computer vision and pattern recognition}, pages 19876--19887, 2024.

\bibitem[Li et~al.(2021)Li, Ghorbani, Broom{\'e}, Rashid, Black, Hernlund, Kjellstr{\"o}m, and Zuffi]{li2021hsmal}
Ci Li, Nima Ghorbani, Sofia Broom{\'e}, Maheen Rashid, Michael~J Black, Elin Hernlund, Hedvig Kjellstr{\"o}m, and Silvia Zuffi.
\newblock hsmal: Detailed horse shape and pose reconstruction for motion pattern recognition.
\newblock \emph{arXiv preprint arXiv:2106.10102}, 2021.

\bibitem[Li et~al.(2024{\natexlab{a}})Li, Mellbin, Krogager, Polikovsky, Holmberg, Ghorbani, Black, Kjellstr{\"o}m, Zuffi, and Hernlund]{li2024poses}
Ci Li, Ylva Mellbin, Johanna Krogager, Senya Polikovsky, Martin Holmberg, Nima Ghorbani, Michael~J Black, Hedvig Kjellstr{\"o}m, Silvia Zuffi, and Elin Hernlund.
\newblock The poses for equine research dataset (pferd).
\newblock \emph{Scientific Data}, 11\penalty0 (1):\penalty0 497, 2024{\natexlab{a}}.

\bibitem[Li et~al.(2024{\natexlab{b}})Li, Yang, Weng, Hernlund, Zuffi, and Kjellstr{\"o}m]{li2024dessie}
Ci Li, Yi Yang, Zehang Weng, Elin Hernlund, Silvia Zuffi, and Hedvig Kjellstr{\"o}m.
\newblock Dessie: Disentanglement for articulated 3d horse shape and pose estimation from images.
\newblock In \emph{Proceedings of the Asian Conference on Computer Vision}, pages 764--783, 2024{\natexlab{b}}.

\bibitem[Li et~al.(2024{\natexlab{c}})Li, Litvak, Li, Zhang, Jakab, Rupprecht, Wu, Vedaldi, and Wu]{li2024learning}
Zizhang Li, Dor Litvak, Ruining Li, Yunzhi Zhang, Tomas Jakab, Christian Rupprecht, Shangzhe Wu, Andrea Vedaldi, and Jiajun Wu.
\newblock Learning the 3d fauna of the web.
\newblock In \emph{Proceedings of the IEEE/CVF Conference on Computer Vision and Pattern Recognition}, pages 9752--9762, 2024{\natexlab{c}}.

\bibitem[Li et~al.(2024{\natexlab{d}})Li, Xing, Fang, Zhang, Fan, and Xu]{li2024multi}
Zhipeng Li, Xiaofen Xing, Yuanbo Fang, Weibin Zhang, Hengsheng Fan, and Xiangmin Xu.
\newblock Multi-scale temporal transformer for speech emotion recognition.
\newblock \emph{arXiv preprint arXiv:2410.00390}, 2024{\natexlab{d}}.

\bibitem[Liang et~al.(2025)Liang, Luo, Chen, Chen, Yan, Li, Liu, and Tan]{liang2025unitex}
Yixun Liang, Kunming Luo, Xiao Chen, Rui Chen, Hongyu Yan, Weiyu Li, Jiarui Liu, and Ping Tan.
\newblock Unitex: Universal high fidelity generative texturing for 3d shapes.
\newblock \emph{arXiv preprint arXiv:2505.23253}, 2025.

\bibitem[Lin et~al.(2014)Lin, Maire, Belongie, Hays, Perona, Ramanan, Doll{\'a}r, and Zitnick]{lin2014microsoft}
Tsung-Yi Lin, Michael Maire, Serge Belongie, James Hays, Pietro Perona, Deva Ramanan, Piotr Doll{\'a}r, and C~Lawrence Zitnick.
\newblock Microsoft coco: Common objects in context.
\newblock In \emph{European conference on computer vision}, pages 740--755. Springer, 2014.

\bibitem[Loshchilov and Hutter(2019)]{adamw}
Ilya Loshchilov and Frank Hutter.
\newblock Decoupled weight decay regularization.
\newblock In \emph{The Seventh International Conference on Learning Representations}, 2019.

\bibitem[Lyu et~al.(2025)Lyu, Zhu, Gu, Lin, Cheng, Liu, Tang, and An]{lyu2025animer}
Jin Lyu, Tianyi Zhu, Yi Gu, Li Lin, Pujin Cheng, Yebin Liu, Xiaoying Tang, and Liang An.
\newblock Animer: Animal pose and shape estimation using family aware transformer.
\newblock In \emph{Proceedings of the Computer Vision and Pattern Recognition Conference}, pages 17486--17496, 2025.

\bibitem[Niewiadomski et~al.(2025)Niewiadomski, Yiannakidis, Cuevas-Velasquez, Sanyal, Black, Zuffi, and Kulits]{niewiadomski2025ICCV}
Tomasz Niewiadomski, Anastasios Yiannakidis, Hanz Cuevas-Velasquez, Soubhik Sanyal, Michael~J. Black, Silvia Zuffi, and Peter Kulits.
\newblock Generative zoo.
\newblock In \emph{Proceedings of the IEEE/CVF International Conference on Computer Vision (ICCV)}, 2025.

\bibitem[Park et~al.(2021)Park, Sinha, Barron, Bouaziz, Goldman, Seitz, and Martin-Brualla]{park2021nerfies}
Keunhong Park, Utkarsh Sinha, Jonathan~T Barron, Sofien Bouaziz, Dan~B Goldman, Steven~M Seitz, and Ricardo Martin-Brualla.
\newblock Nerfies: Deformable neural radiance fields.
\newblock In \emph{Proceedings of the IEEE/CVF international conference on computer vision}, pages 5865--5874, 2021.

\bibitem[Pont-Tuset et~al.(2017)Pont-Tuset, Perazzi, Caelles, Arbel\'aez, Sorkine-Hornung, and {Van Gool}]{Pont-Tuset_arXiv_2017}
Jordi Pont-Tuset, Federico Perazzi, Sergi Caelles, Pablo Arbel\'aez, Alexander Sorkine-Hornung, and Luc {Van Gool}.
\newblock The 2017 davis challenge on video object segmentation.
\newblock \emph{arXiv:1704.00675}, 2017.

\bibitem[Ravi et~al.(2020)Ravi, Reizenstein, Novotny, Gordon, Lo, Johnson, and Gkioxari]{ravi2020accelerating}
Nikhila Ravi, Jeremy Reizenstein, David Novotny, Taylor Gordon, Wan-Yen Lo, Justin Johnson, and Georgia Gkioxari.
\newblock Accelerating 3d deep learning with pytorch3d.
\newblock \emph{arXiv preprint arXiv:2007.08501}, 2020.

\bibitem[Sabathier et~al.(2024)Sabathier, Mitra, and Novotny]{AnimalAvatars2024}
Remy Sabathier, Niloy~J. Mitra, and David Novotny.
\newblock Animal avatars: Reconstructing animatable 3d animals from casual videos.
\newblock In \emph{Computer Vision – ECCV 2024: 18th European Conference, Milan, Italy, September 29–October 4, 2024, Proceedings, Part LXXIX}, page 270–287, 2024.

\bibitem[Sim{\'e}oni et~al.(2025)Sim{\'e}oni, Vo, Seitzer, Baldassarre, Oquab, Jose, Khalidov, Szafraniec, Yi, Ramamonjisoa, et~al.]{simeoni2025dinov3}
Oriane Sim{\'e}oni, Huy~V Vo, Maximilian Seitzer, Federico Baldassarre, Maxime Oquab, Cijo Jose, Vasil Khalidov, Marc Szafraniec, Seungeun Yi, Micha{\"e}l Ramamonjisoa, et~al.
\newblock Dinov3.
\newblock \emph{arXiv preprint arXiv:2508.10104}, 2025.

\bibitem[Sun et~al.(2024)Sun, Litvak, Zhang, Li, Wu, and Wu]{sun2024ponymation}
Keqiang Sun, Dor Litvak, Yunzhi Zhang, Hongsheng Li, Jiajun Wu, and Shangzhe Wu.
\newblock Ponymation: Learning articulated 3d animal motions from unlabeled online videos.
\newblock In \emph{European Conference on Computer Vision}, pages 100--119. Springer, 2024.

\bibitem[Vaswani et~al.(2017)Vaswani, Shazeer, Parmar, Uszkoreit, Jones, Gomez, Kaiser, and Polosukhin]{Vaswani2017AttentionIA}
Ashish Vaswani, Noam~M. Shazeer, Niki Parmar, Jakob Uszkoreit, Llion Jones, Aidan~N. Gomez, Lukasz Kaiser, and Illia Polosukhin.
\newblock Attention is all you need.
\newblock In \emph{Neural Information Processing Systems}, 2017.

\bibitem[Wang et~al.(2025{\natexlab{a}})Wang, Chen, Karaev, Vedaldi, Rupprecht, and Novotny]{wang2025vggt}
Jianyuan Wang, Minghao Chen, Nikita Karaev, Andrea Vedaldi, Christian Rupprecht, and David Novotny.
\newblock Vggt: Visual geometry grounded transformer.
\newblock In \emph{Proceedings of the IEEE/CVF Conference on Computer Vision and Pattern Recognition}, 2025{\natexlab{a}}.

\bibitem[Wang et~al.(2023)Wang, Chang, Cai, Li, Hariharan, Holynski, and Snavely]{wang2023tracking}
Qianqian Wang, Yen-Yu Chang, Ruojin Cai, Zhengqi Li, Bharath Hariharan, Aleksander Holynski, and Noah Snavely.
\newblock Tracking everything everywhere all at once.
\newblock In \emph{Proceedings of the IEEE/CVF International Conference on Computer Vision}, pages 19795--19806, 2023.

\bibitem[Wang et~al.(2025{\natexlab{b}})Wang, Ye, Gao, Zeng, Austin, Li, and Kanazawa]{wang2025shape}
Qianqian Wang, Vickie Ye, Hang Gao, Weijia Zeng, Jake Austin, Zhengqi Li, and Angjoo Kanazawa.
\newblock Shape of motion: 4d reconstruction from a single video.
\newblock In \emph{Proceedings of the IEEE/CVF International Conference on Computer Vision}, pages 9660--9672, 2025{\natexlab{b}}.

\bibitem[Wang et~al.(2024)Wang, Wang, Liu, and Daniilidis]{wang2024tram}
Yufu Wang, Ziyun Wang, Lingjie Liu, and Kostas Daniilidis.
\newblock Tram: Global trajectory and motion of 3d humans from in-the-wild videos.
\newblock In \emph{European Conference on Computer Vision}, pages 467--487. Springer, 2024.

\bibitem[Wu et~al.(2025{\natexlab{a}})Wu, Li, Zhou, Lin, Gao, Yan, Yin, Bai, Xu, Chen, et~al.]{wu2025qwen}
Chenfei Wu, Jiahao Li, Jingren Zhou, Junyang Lin, Kaiyuan Gao, Kun Yan, Sheng-ming Yin, Shuai Bai, Xiao Xu, Yilei Chen, et~al.
\newblock Qwen-image technical report.
\newblock \emph{arXiv preprint arXiv:2508.02324}, 2025{\natexlab{a}}.

\bibitem[Wu et~al.(2025{\natexlab{b}})Wu, Liu, Hung, Qian, Zhan, and Duan]{wu20254d}
Diankun Wu, Fangfu Liu, Yi-Hsin Hung, Yue Qian, Xiaohang Zhan, and Yueqi Duan.
\newblock 4d-fly: Fast 4d reconstruction from a single monocular video.
\newblock In \emph{Proceedings of the Computer Vision and Pattern Recognition Conference}, pages 16663--16673, 2025{\natexlab{b}}.

\bibitem[Wu et~al.(2024)Wu, Yi, Fang, Xie, Zhang, Wei, Liu, Tian, and Wang]{wu20244d}
Guanjun Wu, Taoran Yi, Jiemin Fang, Lingxi Xie, Xiaopeng Zhang, Wei Wei, Wenyu Liu, Qi Tian, and Xinggang Wang.
\newblock 4d gaussian splatting for real-time dynamic scene rendering.
\newblock In \emph{Proceedings of the IEEE/CVF conference on computer vision and pattern recognition}, pages 20310--20320, 2024.

\bibitem[Wu et~al.(2023{\natexlab{a}})Wu, Jakab, Rupprecht, and Vedaldi]{wu2023dove}
Shangzhe Wu, Tomas Jakab, Christian Rupprecht, and Andrea Vedaldi.
\newblock Dove: Learning deformable 3d objects by watching videos.
\newblock \emph{International Journal of Computer Vision}, 131\penalty0 (10):\penalty0 2623--2634, 2023{\natexlab{a}}.

\bibitem[Wu et~al.(2023{\natexlab{b}})Wu, Li, Jakab, Rupprecht, and Vedaldi]{wu2023magicpony}
Shangzhe Wu, Ruining Li, Tomas Jakab, Christian Rupprecht, and Andrea Vedaldi.
\newblock Magicpony: Learning articulated 3d animals in the wild.
\newblock In \emph{Proceedings of the IEEE/CVF Conference on Computer Vision and Pattern Recognition}, pages 8792--8802, 2023{\natexlab{b}}.

\bibitem[Xu et~al.(2023)Xu, Zhang, Zhang, and Tao]{xu2023vitpose++}
Yufei Xu, Jing Zhang, Qiming Zhang, and Dacheng Tao.
\newblock Vitpose++: Vision transformer for generic body pose estimation.
\newblock \emph{IEEE Transactions on Pattern Analysis and Machine Intelligence}, 46\penalty0 (2):\penalty0 1212--1230, 2023.

\bibitem[Yan et~al.(2023)Yan, Hafner, James, and Abbeel]{yan2023temporally}
Wilson Yan, Danijar Hafner, Stephen James, and Pieter Abbeel.
\newblock Temporally consistent transformers for video generation.
\newblock In \emph{International Conference on Machine Learning}, pages 39062--39098. PMLR, 2023.

\bibitem[Yang et~al.(2024)Yang, Huang, Chai, Jiang, and Hwang]{yang2024samurai}
Cheng-Yen Yang, Hsiang-Wei Huang, Wenhao Chai, Zhongyu Jiang, and Jenq-Neng Hwang.
\newblock Samurai: Adapting segment anything model for zero-shot visual tracking with motion-aware memory, 2024.

\bibitem[Yang et~al.(2021{\natexlab{a}})Yang, Sun, Jampani, Vlasic, Cole, Chang, Ramanan, Freeman, and Liu]{yang2021lasr}
Gengshan Yang, Deqing Sun, Varun Jampani, Daniel Vlasic, Forrester Cole, Huiwen Chang, Deva Ramanan, William~T Freeman, and Ce Liu.
\newblock Lasr: Learning articulated shape reconstruction from a monocular video.
\newblock In \emph{Proceedings of the IEEE/CVF Conference on Computer Vision and Pattern Recognition}, pages 15980--15989, 2021{\natexlab{a}}.

\bibitem[Yang et~al.(2021{\natexlab{b}})Yang, Sun, Jampani, Vlasic, Cole, Liu, and Ramanan]{yang2021viser}
Gengshan Yang, Deqing Sun, Varun Jampani, Daniel Vlasic, Forrester Cole, Ce Liu, and Deva Ramanan.
\newblock Viser: Video-specific surface embeddings for articulated 3d shape reconstruction.
\newblock \emph{Advances in Neural Information Processing Systems}, 34:\penalty0 19326--19338, 2021{\natexlab{b}}.

\bibitem[Yang et~al.(2022{\natexlab{a}})Yang, Vo, Neverova, Ramanan, Vedaldi, and Joo]{yang2022banmo}
Gengshan Yang, Minh Vo, Natalia Neverova, Deva Ramanan, Andrea Vedaldi, and Hanbyul Joo.
\newblock Banmo: Building animatable 3d neural models from many casual videos.
\newblock In \emph{Proceedings of the IEEE/CVF Conference on Computer Vision and Pattern Recognition}, pages 2863--2873, 2022{\natexlab{a}}.

\bibitem[Yang et~al.(2023)Yang, Wang, Reddy, and Ramanan]{yang2023reconstructing}
Gengshan Yang, Chaoyang Wang, N~Dinesh Reddy, and Deva Ramanan.
\newblock Reconstructing animatable categories from videos.
\newblock In \emph{Proceedings of the IEEE/CVF Conference on Computer Vision and Pattern Recognition}, pages 16995--17005, 2023.

\bibitem[Yang et~al.(2022{\natexlab{b}})Yang, Yang, Xu, Zhang, Lan, and Tao]{yang2022apt}
Yuxiang Yang, Junjie Yang, Yufei Xu, Jing Zhang, Long Lan, and Dacheng Tao.
\newblock Apt-36k: A large-scale benchmark for animal pose estimation and tracking.
\newblock \emph{Advances in Neural Information Processing Systems}, 35:\penalty0 17301--17313, 2022{\natexlab{b}}.

\bibitem[Yao et~al.(2022)Yao, Hung, Li, Rubinstein, Yang, and Jampani]{yao2022lassie}
Chun-Han Yao, Wei-Chih Hung, Yuanzhen Li, Michael Rubinstein, Ming-Hsuan Yang, and Varun Jampani.
\newblock Lassie: Learning articulated shapes from sparse image ensemble via 3d part discovery.
\newblock \emph{Advances in Neural Information Processing Systems}, 35:\penalty0 15296--15308, 2022.

\bibitem[Yao et~al.(2023)Yao, Hung, Li, Rubinstein, Yang, and Jampani]{yao2023hi}
Chun-Han Yao, Wei-Chih Hung, Yuanzhen Li, Michael Rubinstein, Ming-Hsuan Yang, and Varun Jampani.
\newblock Hi-lassie: High-fidelity articulated shape and skeleton discovery from sparse image ensemble.
\newblock In \emph{Proceedings of the IEEE/CVF Conference on Computer Vision and Pattern Recognition}, pages 4853--4862, 2023.

\bibitem[Zhang et~al.(2025)Zhang, Xu, Wang, Yang, Zhao, Chen, and Guo]{zhang2025gaussian}
Bowen Zhang, Sicheng Xu, Chuxin Wang, Jiaolong Yang, Feng Zhao, Dong Chen, and Baining Guo.
\newblock Gaussian variation field diffusion for high-fidelity video-to-4d synthesis.
\newblock \emph{arXiv preprint arXiv:2507.23785}, 2025.

\bibitem[Zhang et~al.(2024)Zhang, Herrmann, Hur, Jampani, Darrell, Cole, Sun, and Yang]{zhang2024monst3r}
Junyi Zhang, Charles Herrmann, Junhwa Hur, Varun Jampani, Trevor Darrell, Forrester Cole, Deqing Sun, and Ming-Hsuan Yang.
\newblock Monst3r: A simple approach for estimating geometry in the presence of motion.
\newblock \emph{arXiv preprint arXiv:2410.03825}, 2024.

\bibitem[Zhang et~al.(2018)Zhang, Isola, Efros, Shechtman, and Wang]{zhang2018unreasonable}
Richard Zhang, Phillip Isola, Alexei~A Efros, Eli Shechtman, and Oliver Wang.
\newblock The unreasonable effectiveness of deep features as a perceptual metric.
\newblock In \emph{Proceedings of the IEEE conference on computer vision and pattern recognition}, pages 586--595, 2018.

\bibitem[Zhao et~al.(2025)Zhao, Wu, and Wu]{Animal4D}
Brian~Nlong Zhao, Jiajun Wu, and Shangzhe Wu.
\newblock Web-scale collection of video data for 4d animal reconstruction, 2025.

\bibitem[Zhao et~al.(2021)Zhao, Jiang, Jia, Torr, and Koltun]{zhao2021point}
Hengshuang Zhao, Li Jiang, Jiaya Jia, Philip~HS Torr, and Vladlen Koltun.
\newblock Point transformer.
\newblock In \emph{Proceedings of the IEEE/CVF international conference on computer vision}, pages 16259--16268, 2021.

\bibitem[Zhong et~al.(2025)Zhong, Peng, Zheng, Huang, Ma, Zhang, Liu, Yuille, and Chen]{zhong20254d}
Shanshan Zhong, Jiawei Peng, Zehan Zheng, Zhongzhan Huang, Wufei Ma, Guofeng Zhang, Qihao Liu, Alan Yuille, and Jieneng Chen.
\newblock 4d-animal: Freely reconstructing animatable 3d animals from videos.
\newblock \emph{arXiv preprint arXiv:2507.10437}, 2025.

\bibitem[Zhou et~al.(2025)Zhou, Wang, Chen, Chang, Beaudouin, Zhan, Liang, and Wang]{zhou2025page}
Kaichen Zhou, Yuhan Wang, Grace Chen, Xinhai Chang, Gaspard Beaudouin, Fangneng Zhan, Paul~Pu Liang, and Mengyu Wang.
\newblock Page-4d: Disentangled pose and geometry estimation for 4d perception.
\newblock \emph{arXiv preprint arXiv:2510.17568}, 2025.

\bibitem[Zuffi and Black(2024)]{zuffi_eccv2024_awol}
Silvia Zuffi and Michael~J. Black.
\newblock Awol: Analysis without synthesis using language.
\newblock In \emph{European Conference on Computer Vision (ECCV)}, 2024.

\bibitem[Zuffi et~al.(2017)Zuffi, Kanazawa, Jacobs, and Black]{Zuffi:CVPR:2017}
Silvia Zuffi, Angjoo Kanazawa, David Jacobs, and Michael~J. Black.
\newblock {3D} menagerie: Modeling the {3D} shape and pose of animals.
\newblock In \emph{IEEE Conf. on Computer Vision and Pattern Recognition (CVPR)}, 2017.

\bibitem[Zuffi et~al.(2018)Zuffi, Kanazawa, and Black]{zuffi2018lions}
Silvia Zuffi, Angjoo Kanazawa, and Michael~J Black.
\newblock Lions and tigers and bears: Capturing non-rigid, 3d, articulated shape from images.
\newblock In \emph{Proceedings of the IEEE conference on Computer Vision and Pattern Recognition}, pages 3955--3963, 2018.

\bibitem[Zuffi et~al.(2019)Zuffi, Kanazawa, Berger-Wolf, and Black]{Zuffi:ICCV:2019}
Silvia Zuffi, Angjoo Kanazawa, Tanya Berger-Wolf, and Michael~J. Black.
\newblock Three-d safari: Learning to estimate zebra pose, shape, and texture from images "in the wild".
\newblock In \emph{International Conference on Computer Vision}, 2019.

\bibitem[Zuffi et~al.(2024)Zuffi, Mellbin, Li, Hoeschle, Kjellström, Polikovsky, Hernlund, and Black]{Zuffi:CVPR:2024}
Silvia Zuffi, Ylva Mellbin, Ci Li, Markus Hoeschle, Hedvig Kjellström, Senya Polikovsky, Elin Hernlund, and Michael~J. Black.
\newblock {VAREN}: Very accurate and realistic equine network.
\newblock In \emph{IEEE/CVF Conference on Computer Vision and Pattern Recognition (CVPR)}, 2024.

\end{thebibliography}
